# An Evolutionary Algorithm with Advanced Goal and Priority Specification for Multi-objective Optimization

**Kay Chen Tan** ELETANKC@NUS.EDU.SG
**Eik Fun Khor** EIKFUN.KHOR@SEAGATE.COM
**Tong Heng Lee** ELELEETH@NUS.EDU.SG
**Ramasubramanian Sathikannan** K.SATHI@GSK.COM
*National University of Singapore*
*4 Engineering Drive 3, Singapore 117576*
*Republic of Singapore*

**Abstract**

This paper presents an evolutionary algorithm with a new goal-sequence domination scheme for better decision support in multi-objective optimization. The approach allows the inclusion of advanced hard/soft priority and constraint information on each objective component, and is capable of incorporating multiple specifications with overlapping or non-overlapping objective functions via logical "OR" and "AND" connectives to drive the search towards multiple regions of trade-off. In addition, we propose a dynamic sharing scheme that is simple and adaptively estimated according to the on-line population distribution without needing any a priori parameter setting. Each feature in the proposed algorithm is examined to show its respective contribution, and the performance of the algorithm is compared with other evolutionary optimization methods. It is shown that the proposed algorithm has performed well in the diversity of evolutionary search and uniform distribution of non-dominated individuals along the final trade-offs, without significant computational effort. The algorithm is also applied to the design optimization of a practical servo control system for hard disk drives with a single voice-coil-motor actuator. Results of the evolutionary designed servo control system show a superior closed-loop performance compared to classical PID or RPT approaches.

## 1. Introduction

Many real-world design tasks involve optimizing a vector of objective functions on a feasible decision variable space. These objective functions are often non-commensurable and in competition with each other, and cannot be simply aggregated into a scalar function for optimization. This type of problem is known as multi-objective (MO) optimization problem, for which the solution is a family of points known as a Pareto-optimal set (Goldberg, 1989), where each objective component of any member in the set can only be improved by degrading at least one of its other objective components. To obtain a good solution via conventional MO optimization techniques such as the methods of inequalities, goal attainment or weighted sum approach, a continuous cost function and/or a set of precise settings of weights or goals are required, which are usually not well manageable or understood (Grace, 1992; Osyczka, 1984).

Emulating the Darwinian-Wallace principle of "survival-of-the-fittest" in natural selection and genetics, evolutionary algorithms (EAs) (Holland, 1975) have been found to be effective and efficient in solving complex problems where conventional optimization tools fail to work well.

The EAs evaluate performances of candidate solutions at multiple points simultaneously, and are capable of approaching the global optimum in a noisy, poorly understood and/or non-differentiable search space (Goldberg, 1989).

Since Schaffer's work (1985), evolutionary algorithm-based search techniques for MO optimization have been gaining significant attention from researchers in various disciplines. This is reflected by the high volume of publications in this topic in the last few years as well as the first international conference on Evolutionary Multi-criteria Optimization (EMO'01) held in March 2001 at Zurich, Switzerland. Readers may refer to (Coello Coello, 1996; 1999; Deb, 2001; Fonseca, 1995; Van Veldhuizen & Lamont, 1998; Zitzler & Thiele, 1999) on detailed implementation of various evolutionary techniques for MO optimization.

Unlike most conventional methods that linearly combine multiple attributes to form a composite scalar objective function, the concept of Pareto's optimality or modified selection scheme is incorporated in an evolutionary MO optimization to evolve a family of solutions at multiple points along the trade-off surface simultaneously (Fonseca & Fleming, 1993). Among various selection techniques for evolutionary MO optimization, the Pareto-dominance scheme (Goldberg, 1989) that assigns equal rank to all non-dominated individuals is an effective approach for comparing the strengths among different candidate solutions in a population (Fonseca & Fleming, 1993). Starting from this principle, Fonseca and Fleming (1993) proposed a Pareto-based ranking scheme to include goal and priority information for MO optimization. The underlying reason is that certain user knowledge may be available for an optimization problem, such as preferences and/or goals to be achieved for certain objective components. This information could be useful and incorporated by means of goal and priority vectors, which simplify the optimization process and allow the evolution to be directed towards certain concentrated regions of the trade-offs. Although the ranking scheme is a good approach, it only works for a single goal and priority vector setting, which may be difficult to define accurately prior to an optimization process for real-world optimization problems. Moreover, the scheme does not allow advanced specifications, such as logical "AND" and "OR" operations among multiple goals and priorities.

Based on the Pareto-based domination approach, this paper reformulates the domination scheme to incorporate advanced specifications for better decision support in MO optimization. Besides the flexibility of incorporating goal and priority information on each objective component, the proposed domination scheme allows the inclusion of hard/soft priority and constraint specifications. In addition, the approach is capable of incorporating multiple specifications with overlapping or non-overlapping objective functions via logical "OR" and "AND" connectives to drive the search towards multiple regions of the trade-off. The paper also proposes a dynamic sharing scheme, which computes the sharing distance adaptively based upon the on-line population distribution in the objective domain without the need of any a priori parameter setting. The dynamic sharing approach is essential since it eliminates the difficulty of manually finding an appropriate sharing distance prior to an optimization process. The choice of such a distance would be sensitive to the size and geometry of the discovered trade-offs (Coello Coello, 1999; Fonseca & Fleming, 1993).

This paper is organized as follows: The formulation of the proposed domination scheme for better decision support is presented in Section 2. A dynamic sharing scheme that estimates the sharing distance adaptively based upon the on-line population distribution is described in Section 3. Section 4 examines the usefulness and contribution of each proposed feature in the algorithm. The performance comparison of the proposed algorithm with other evolutionary MO optimization methods is also shown in the section. Practical application of the proposed algorithm to servo control system design optimization is given in Section 5. Conclusions are drawn in Section 6.

## 2. Advanced Goal and Priority Specifications for MO Optimization

A multi-objective optimization problem seeks to optimize a vector of non-commensurable and often competing objectives, i.e., it tends to find a parameter set $\boldsymbol{P}$ for $\underset{\boldsymbol{P}\in\Phi}{\text{Min}}\,\boldsymbol{F}(\boldsymbol{P})$, $\boldsymbol{P} \in R^n$, where $\boldsymbol{P} = \{p_1, p_2, \ldots, p_n\}$ is a $n$-dimensional vector having $n$ decision variables or parameters, and $\Phi$ defines a feasible set of $\boldsymbol{P}$. $\boldsymbol{F} = \{f_1, f_2, \ldots, f_m\}$ is an objective vector with $m$ objectives to be minimized. For a MO optimization problem with simple goal or priority specification on each objective function, the Pareto-based ranking scheme is sufficient (Fonseca & Fleming, 1993). In practice, however, it may be difficult to define an accurate goal and priority setting in a priori to an optimization process for real-world optimization problems. Besides goal and priority information, there could also be additional supporting specifications that are useful or need to be satisfied in the evolutionary search, such as optimization constraints or feasibility of a solution. Moreover, the Pareto-based ranking scheme does not allow advanced specifications, such as logical "AND" and "OR" operations among multiple goals and priorities for better decision support in complex MO optimization. In this section, a new goal-sequence Pareto-based domination scheme is proposed to address these issues and to provide hard/soft goal and priority specifications for better controls in the evolutionary optimization process.

### 2.1 Pareto-based Domination with Goal Information

This section is about an effective two-stage Pareto-based domination scheme for MO optimization, which is then extended to incorporate advanced soft/hard goal and priority specifications. Consider a minimization problem. An objective vector $\boldsymbol{F}_a$ is said to dominate another objective vector $\boldsymbol{F}_b$ based on the idea of Pareto dominance, denoted by $\boldsymbol{F}_a \;\pi\; \boldsymbol{F}_b$, *iff* $f_{a,i} \le f_{b,i}\ \forall\ i \in \{1,2,..., m\}$ and $f_{a,j} < f_{b,j}$ for some $j \in \{1,2,..., m\}$. Adopting this basic principle of Pareto dominance, the first stage in the proposed domination approach ranks all individuals that satisfy the goal setting to minimize the objective functions as much as possible. It assigns the same smallest cost for all non-dominated individuals, while the dominated individuals are ranked according to how many individuals in the population dominate them. The second stage ranks the remaining individuals that do not meet the goal setting based upon the following extended domination scheme. Let $\boldsymbol{F}_a^{\breve{a}}$ and $\boldsymbol{F}_b^{\breve{a}}$ denote the component of vector $\boldsymbol{F}_a$ and $\boldsymbol{F}_b$ respectively, in which $\boldsymbol{F}_a$ does not meet the goal $\boldsymbol{G}$. Then for both $\boldsymbol{F}_a$ and $\boldsymbol{F}_b$ that do not totally satisfy the goal $\boldsymbol{G}$, the vector $\boldsymbol{F}_a$ is said to dominate vector $\boldsymbol{F}_b$ (denoted by $\boldsymbol{F}_a \underset{\boldsymbol{G}}{\pi} \boldsymbol{F}_b$) *iff*

$$(\boldsymbol{F}_a^{\}a} \;\pi\; \boldsymbol{F}_b^{\}a}) \text{ or } (abs(\boldsymbol{F}_a\text{-}\boldsymbol{G}) \;\pi\; abs(\boldsymbol{F}_b\text{-}\boldsymbol{G})) \quad (1)$$

For this, the rank begins from one increment of the maximum rank value obtained in the first stage of the cost assignment. Therefore individuals that do not meet the goal will be directed toward the goal and the infinum in the objective domain, while those that have satisfied the goal will only be directed further towards the infinum. Note that the domination comparison operator is non-commutative ($\boldsymbol{F}_a \underset{\boldsymbol{G}}{\pi} \boldsymbol{F}_b \neq \boldsymbol{F}_b \underset{\boldsymbol{G}}{\pi} \boldsymbol{F}_a$). Figure 1 shows an optimization problem with two objectives $f_1$ and $f_2$ to be minimized. The arrows in Figure 1 indicate the transformation according to $\boldsymbol{F}' = |\boldsymbol{F} - \boldsymbol{G}|$ of the objective function $\boldsymbol{F}$ to $\boldsymbol{F}'$ for individuals that do not satisfy the goal, with the goal as the new reference point in the transformed objective domain. It is obvious that the domination scheme is simple and efficient for comparing the strengths among partially or totally unsatisfactory individuals in a population. For comparisons among totally satisfactory individuals, the basic Pareto-dominance is sufficient.

To study the computational efficiency in the approach, the population is divided into two separate groups classified by the goal satisfaction, and the domination comparison is performed separately in each group of individuals. The total number of domination comparisons for the two-stage domination scheme is $N_c$ = $[n_G^{\{}(n_G^{\{}\text{-}1) + n_G^{\}}(n_G^{\}}\text{-}1)]$ where $n_G^{\{}$ is the number of individuals that completely satisfy the goal $\boldsymbol{G}$ and $n_G^{\}}$ is the number of individuals partially satisfy or completely not satisfy the goal $\boldsymbol{G}$. Note that $n_G^{\{} + n_G^{\}} = N$ for a population size of $N$. Hence, in any generation, $N_c$ is always less than or equal to the total number of domination comparisons among all individuals in a population (each individual in the population is compared with ($N$-1) individuals), i.e., $N_c \leq N_{nc} = N(N-1)$. In the next section, the two-stage Pareto-based domination scheme will be extended to incorporate soft/hard priority specifications for advanced MO optimization.

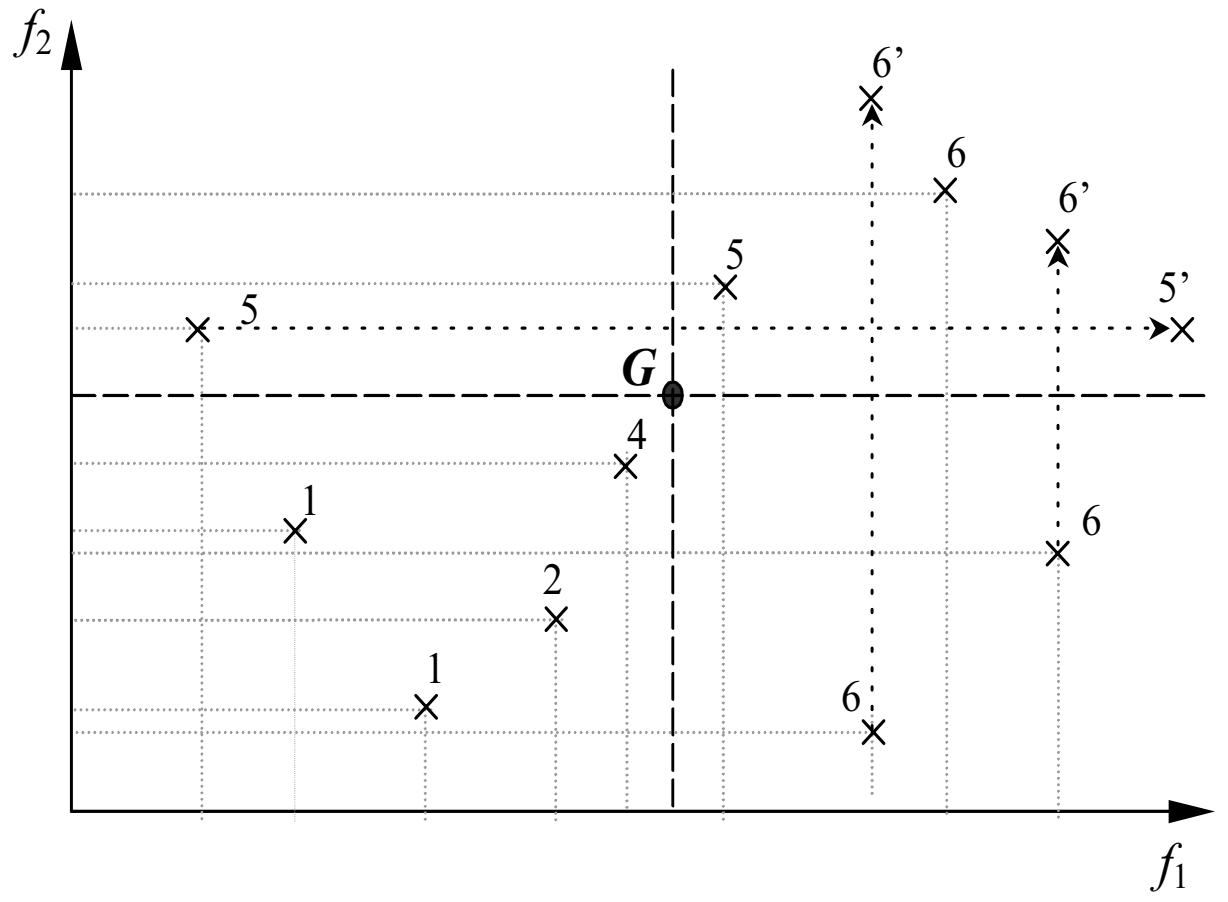

Figure 1: Advanced Pareto Domination Scheme with Goal Information

### 2.2 Goal-Sequence Domination Scheme with Soft/Hard Priority Specifications

One of the advanced capabilities in evolutionary MO optimization is to incorporate cognitive specification, such as priority information that indicates the relative importance of the multiple tasks to provide useful guidance in the optimization. Consider a problem with multiple non-commensurable tasks, where each task is assigned a qualitative form of priority indicating its relative importance. In general, there exist two alternatives to accomplish these tasks, i.e., to consider one task at a time in a sequence according to the task priority or to accomplish all tasks at once before considering any individual task according to the task priority. Intuitively, the former approach provides good optimization performance for tasks with higher priority and may result in relatively poor performance for others. This is due to the fact that optimizing the higher priority tasks may be at the performance expense of the lower priority tasks. This definition of priority is denoted as “hard” priority in this paper. On the other hand, the latter approach provides a distributed approach in which all tasks aim at a compromise solution before the importance or priority of individual task is considered. This is defined as "soft" priority. Similarly, priorities for different objective components in MO optimization can be classified as "hard" or "soft" priority. With hard priorities, goal settings (if applicable) for higher priority objective components must be satisfied first before attaining goals with lower priority. In contrast, soft priorities will first optimize the overall performance of all objective components, as much as possible, before attaining any goal setting of an individual objective component in a sequence according to the priority vector.

To achieve greater flexibility in MO optimization, the two-stage Pareto-based domination scheme is further extended to incorporate both soft and hard priority specifications with or without goal information by means of a new goal-sequence domination. Here, instead of having one priority vector to indicate priorities among the multiple objective components (Fonseca & Fleming, 1998), two kinds of priority vectors are used to accommodate the soft/hard priority information. Consider an objective priority vector, $\boldsymbol{P}_f \in \aleph^{1\mathrm{x}m}$ and a goal priority vector, $\boldsymbol{P}_g \in \aleph^{1\mathrm{x}m}$, where $\boldsymbol{P}_f(i)$ represents the priority for the $i^{\text{th}}$ objective component $\boldsymbol{F}(i)$ that is to be minimized; $\boldsymbol{P}_g(i)$ denotes the priority for the $i^{th}$ goal component $\boldsymbol{G}(i)$ that is to be attained; $m$ is the number of objectives to be minimized and $\aleph$ denotes the natural numbers. The elements of the vector $\boldsymbol{P}_f$ and $\boldsymbol{P}_g$ can take any value in the natural numbers, with a lower number representing a higher priority and zero representing a "don't care" priority assignment. Note that repeated values among the elements in $\boldsymbol{P}_f$ and $\boldsymbol{P}_g$ can be used to indicate equal priority provided that $\boldsymbol{P}_f(i) \neq \boldsymbol{P}_g(i)\ \forall\ i \in \{1, 2, \ldots, m\}$, avoiding contradiction of the priority assignment. With the combination of an objective priority vector $\boldsymbol{P}_f$ and a goal priority vector $\boldsymbol{P}_g$, soft and hard priorities can be defined provided that there is more than one preference among the objective components as given by

$$\exists\ \{(\boldsymbol{P}_f : \boldsymbol{P}_f(j) > 1)\ \vee\ (\boldsymbol{P}_g : \boldsymbol{P}_g(j) > 1)\} \text{ for some } j \in \{1, 2, \ldots, m\} \tag{2}$$

Based on this, a priority setting is regarded as "soft" *iff*

$$\forall\ i \in \{1, 2, \ldots, m\}\ \exists\ \{(\boldsymbol{P}_f : \boldsymbol{P}_f(i) = 1)\ \vee\ (\boldsymbol{P}_g : \boldsymbol{P}_g(i) = 1)\} \tag{3}$$

else, the priority is denoted as "hard".

For example, the settings of $\boldsymbol{P}_f = [1, 1, 2, 2]$ and $\boldsymbol{P}_g = [0, 0, 0, 0]$ for a 4-objective optimization problem indicate that the first and second objective components are given top priority to be minimized, as much as possible, before considering minimization of the third and fourth objective components. Since all elements in $\boldsymbol{P}_g$ are zeros (don't care), no goal components will be considered in the minimization in this case. On the other hand, the setting of $\boldsymbol{P}_f = [0, 0, 0, 0]$ and $\boldsymbol{P}_g = [1, 1, 2, 2]$ imply that the first and second objective components are given the first priority to meet their respective goal components before considering the goal attainment for the third and fourth objective components. The above two different priority settings are all categorized as hard priorities since in both cases, objective components with higher priority are minimized before considering objective components with lower priority. For soft priority as defined in Eq. 3, the objective priority vector and goal priority vector can be set as $\boldsymbol{P}_g = [1, 1, 1, 1]$ and $\boldsymbol{P}_f = [2, 2, 3, 3]$, respectively. This implies that the evolution is directed towards minimizing all of the objective components to the goal region before any attempt to minimize the higher priority objective components in a sequence defined by the priority vector.

To systematically rank all individuals in a population to incorporate the soft/hard priority specifications, a sequence of goals corresponding to the priority information can be generated and represented by a goal-sequence matrix $\boldsymbol{G'}$ where the $k^{th}$ row in the matrix represents the goal vector for the corresponding $k^{th}$ priority. The number of goal vectors to be generated depends on the last level of priority $z$, where $z$ is the maximum value of any one element of $\boldsymbol{P}_g$ and $\boldsymbol{P}_f$ as given by

$$z = \max[\boldsymbol{P}_g(i), \boldsymbol{P}_f(j)] \quad \forall\ i, j \in \{1,2,\ldots,m\} \tag{4}$$

For this, the goal vectors with $k^{th}$ priority in the goal-sequence matrix $\boldsymbol{G'}_k(i)$ for the priority index $k = 1, 2,\ldots, z$ is defined as

$$\forall i = 1,\ldots,m,\ \ \boldsymbol{G'}_k(i) = \begin{cases} \boldsymbol{G}(i) & \textit{if } \boldsymbol{P}_g(i) = k \\ \min[\boldsymbol{F}_{j=1,\ldots,N}(i)] & \textit{if } \boldsymbol{P}_f(i) = k \\ \max[\boldsymbol{F}_{j=1,\ldots,N}(i)] & \textit{otherwise} \end{cases} \tag{5}$$

where $N$ denotes the population size; $\min[\boldsymbol{F}_{j=1,\ldots,N}(i)]$ and $\max[\boldsymbol{F}_{j=1,\ldots,N}(i)]$ represents the minimum and maximum value of the $i^{th}$ objective function from the on-line population distribution, respectively. In Eq. 5, for any $i^{th}$ objective component of any $k$ priority level, the reason for assigning $\boldsymbol{G}'_k(i)$ with $\boldsymbol{G}(i)$ is to guide the individuals towards the goal regions; $\min[\boldsymbol{F}_{j=1,\ldots,N}(i)]$ is to minimize the corresponding objective component as much as possible; and $\max[\boldsymbol{F}_{j=1,\ldots,N}(i)]$ is to relax the requirements on the individuals to give other objective components more room for improvement. According to Eq. 5, the goal-sequence matrix $\boldsymbol{G}'_k(i)$ is dynamic at each generation, as the values of $\min[\boldsymbol{F}_{j=1,\ldots,N}(i)]$ and $\max[\boldsymbol{F}_{j=1,\ldots,N}(i)]$ are dynamically computed depending on the on-line population distribution. After computing the sequence of goals $\boldsymbol{G}'_k$ $\forall$ $k \in \{1, 2,\ldots, z\}$, the individuals are first ranked according to the computed goal $\boldsymbol{G}'_1$ for the first priority. Then each group of individuals that has the same ranks will be further compared and ranked according to next goal $\boldsymbol{G}'_2$ for the second priority to further evaluate the individuals' domination in a population. In general, this ranking process continues until there is no individual with the same rank value or after ranking the goal $\boldsymbol{G}'_z$ that has the lowest priority in the goal-sequence matrix. Note that individuals with the same rank value will not be further evaluated for those components with "don't care" assignments.

With the proposed goal-sequence domination scheme as given in Eq. 5, both hard and soft priority specifications can be incorporated in MO optimization. Without loss of generality, consider a two-objective optimization problem, with $f_1$ having a higher priority than $f_2$, as well as a goal setting of $\boldsymbol{G} = [g_1, g_2]$. For soft priority optimization as defined in Eq. 3, the goal priority vector and objective priority vector can be set as $\boldsymbol{P}_g = [1, 1]$ and $\boldsymbol{P}_f = [2, 0]$, respectively. Let $\min[\boldsymbol{F}(i)]$ and $\max[\boldsymbol{F}(i)]$ denote the minimum and maximum value of the $i$-objective component of $\boldsymbol{F}$ in a population, respectively. The relevant goals in the goal-sequence matrix for each priority level as defined in Eq. 5 are then given as $\boldsymbol{G}'_1 = \boldsymbol{G}$ for the first priority and $\boldsymbol{G}'_2 = \{\min[\boldsymbol{F}(1)], \max[\boldsymbol{F}(2)]\}$ for the second priority. The goal-sequence domination scheme for the two-objective minimization problem is illustrated in Figure 2. Here, the rank value of each individual is denoted by $r_1 \rightarrow r_2$, where $r_1$ and $r_2$ is the rank value after the goal-sequence ranking of the first and second priority, respectively. The preference setting indicates that both $g_1$ and $g_2$ are given the same priority to be attained in the optimization before individuals are further ranked according to the higher priority of $f_1$. This is illustrated in Figure 3a, which shows the location of the desired Pareto-front (represented by the dark region) and the expected evolution direction (represented by the curved arrow) in the objective domain for an example with an unfeasible goal setting $\boldsymbol{G}$.

For hard priority optimization as defined in Eqs. 2 and 3, the goal priority vector and objective priority vector can be set as $\boldsymbol{P}_g = [1, 2]$ and $\boldsymbol{P}_f = [0, 0]$, respectively. According to Eq. 5, this gives a goal sequence of $\boldsymbol{G}'_1 = [g_1, \max[\boldsymbol{F}(2)]$ and $\boldsymbol{G}'_2 = [\max[\boldsymbol{F}(1)], g_2]$ for the first and second priority, respectively. It implies that $g_1$ is given higher priority than $g_2$ to be attained in the optimization. Figure 3b shows the location of the desired Pareto-front (represented by dark region) and the expected evolution direction (represented by curved arrow) in the objective domain. As compared to the solutions obtained in soft priority optimization, hard priority optimization attempts to attain the first goal component and leads to the solution with better $f_1$ (higher priority) but worse $f_2$ (lower priority). It should be mentioned that the setting of soft/hard priority may be subjective or problem dependent in practice. In general, the hard priority optimization may be appropriate for problems with well-defined goals in order to avoid stagnation with unfeasible goal settings. Soft priority optimization is more suitable for applications where moderate performance among various objective components is desired. Besides soft/hard priority information, there may be additional specifications such as optimization constraints that are required to be satisfied in the optimization. These specifications could be easily incorporated in MO optimization by formulating the constraints as additional objective components to be optimized (Fonseca & Fleming, 1998). This will be discussed in the next section.

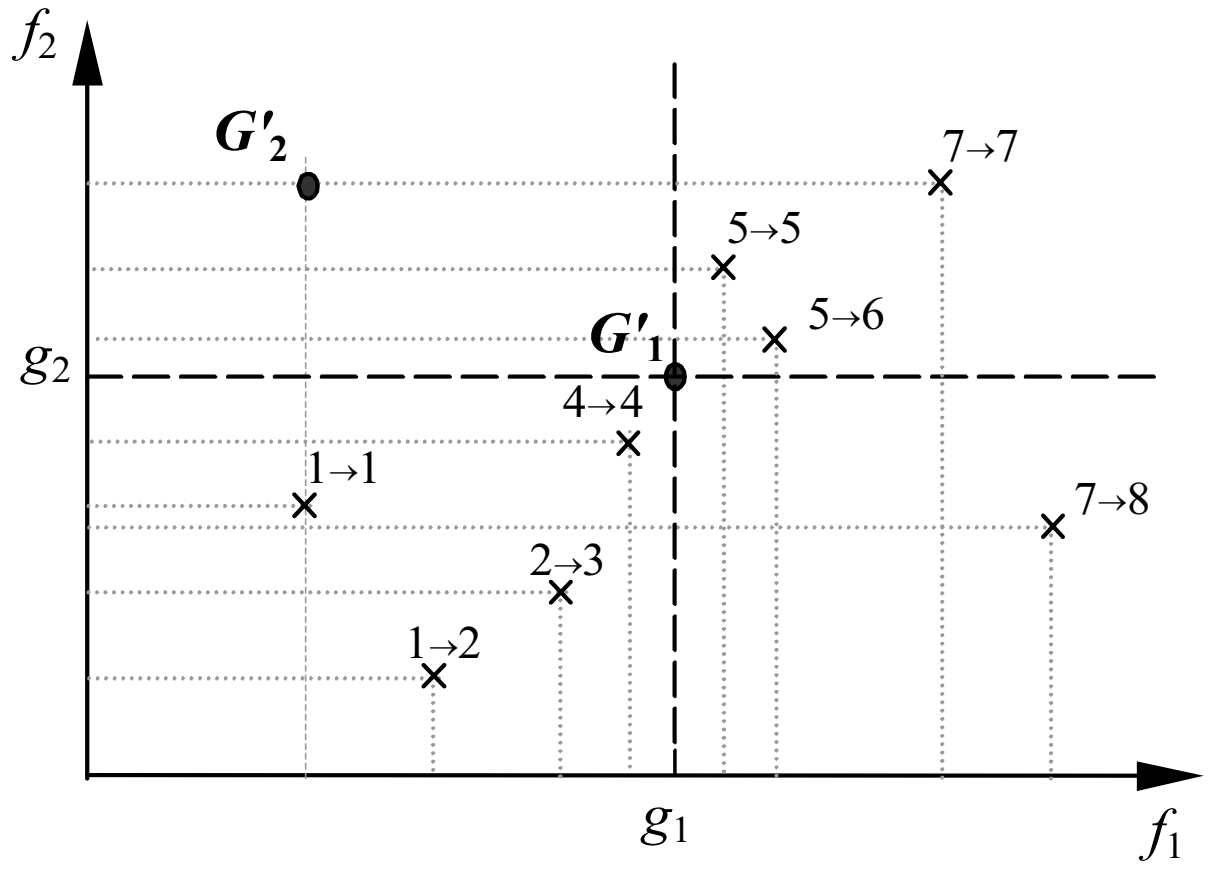

Figure 2: Goal-sequence Domination with Goal $\boldsymbol{G} = \{g_1, g_2\}$, Priority $\boldsymbol{P}_g = [1, 1]$ and $\boldsymbol{P}_f = [2, 0]$

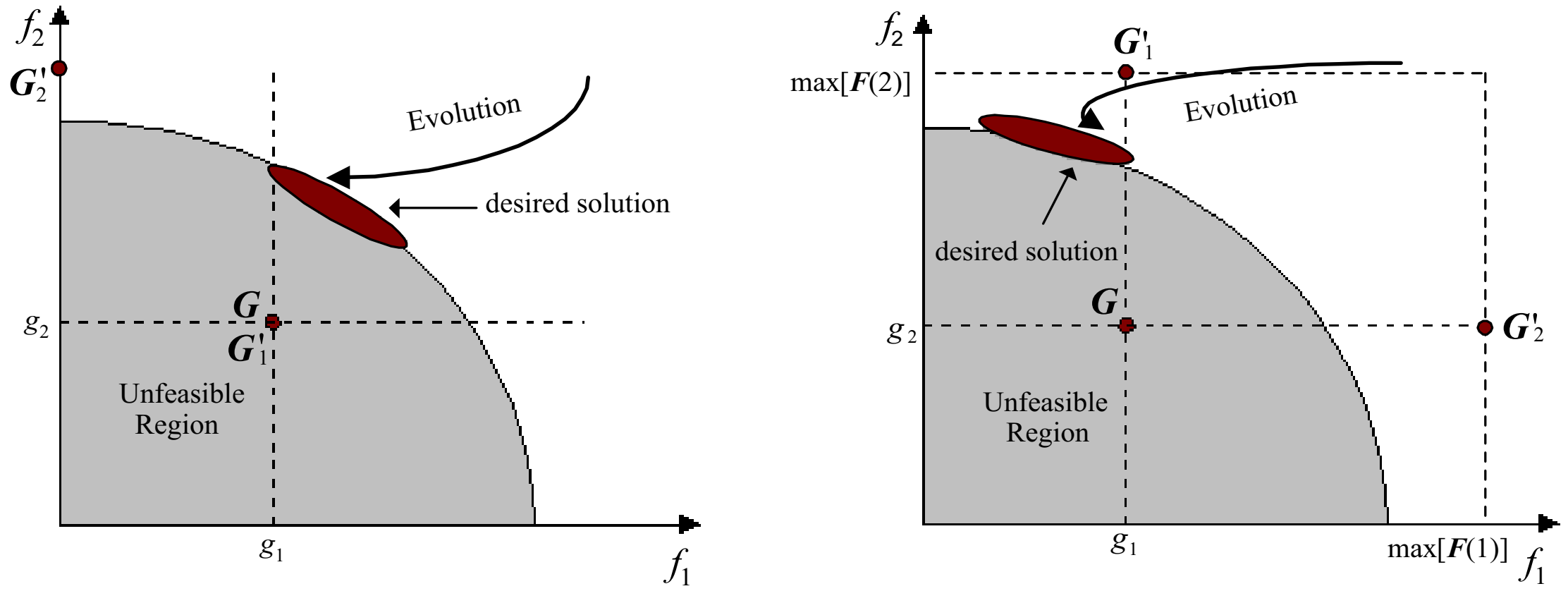

(a) Soft priority $f_1$ higher than $f_2$ (b) Hard priority $f_1$ higher than $f_2$

Figure 3: Illustration of Soft and Hard Priority with Unfeasible Goal Setting

### 2.3 Optimization with Soft and Hard Constraints

Constraints often exist in practical optimization problems (Luus et al. 1995; Michalewicz & Schoenauer, 1996). These constraints are often incorporated in the MO optimization function as one of the objective components to be optimized. It could be in the form of "hard" constraint where the optimization is directed towards attaining a threshold or goal, and further optimization is meaningless or not desirable whenever the goal has been satisfied. In contrast, a "soft" constraint requires that the value of objective component corresponding to the constraint is optimized as much as possible. An easy approach to deal with both hard and soft constraints concurrently in evolutionary MO optimization is given here. At each generation, an updated objective function $\boldsymbol{F}_x^{\#}$ concerning both hard and soft constraints for an individual $x$ with its objective function $\boldsymbol{F}_x$ can be computed in a priori to the goal-sequence domination scheme as given by

$$\boldsymbol{F}_{\boldsymbol{x}}^{\#}(i) = \begin{cases} \boldsymbol{G}(i) & if \ [\boldsymbol{G}(i) \text{ is hard}] \,\&\, [\boldsymbol{F}_{\boldsymbol{x}}(i) < \boldsymbol{G}(i)] \\ \boldsymbol{F}_{\boldsymbol{x}}(i) & otherwise \end{cases} \qquad \forall i = \{1,\ldots,m\} \tag{6}$$

In Eq. 6, any objective component $i$ that corresponds to a hard constraint is assigned to the value of $\boldsymbol{G}(i)$ whenever the hard constraint has been satisfied. The underlying reason is that there is no ranking preference for any particular objective component that has the same value in an evolutionary optimization process, and thus the evolution will only be directed towards optimizing soft constraints and any unattained hard constraints, as desired.

### 2.4 Logical Connectives among Goal and Priority Specifications

For MO optimization problems with a single goal or priority specification, the decision maker often needs to “guess” an appropriate initial goal or priority vector and then manually observe the optimization progress. If any of the goal components is too stringent or too generous, the goal setting will have to be adjusted accordingly until a satisfactory solution can be obtained. This approach obviously requires extensive human observation and intervention, which can be tedious or inefficient in practice. Marcu (1997) proposed a method of adapting the goal values based upon the on-line population distribution at every generation. However, the adaptation of goal values is formulated in such a way that the search is always uniformly directed towards the middle region of the trade-offs. This restriction may be undesirable for many applications, where the trade-off surface is unknown or the search needs to be directed in any direction other than the middle region of the trade-off surface. To reduce human interaction and to allow multiple sets of goal and priority specifications that direct the evolutionary search towards a different portion of the trade-off surface in a single run, the goal-sequence domination scheme is extended in this section to enable logical statements such as “OR” ($\cup$) and “AND” ($\cap$) operations among multiple goal and priority specifications.

These logical operations can be built on top of the goal-sequence domination procedure for each specification. By doing this, the unified rank value for each individual can be determined and taken into effect immediately in the evolution towards the regions concerned. Consider ranking an objective vector $\boldsymbol{F}_x$ by comparing it to the rest of the individuals in a population with reference to two different specification settings of $\boldsymbol{S}_i$ and $\boldsymbol{S}_j$, where $\boldsymbol{S}_i$ and $\boldsymbol{S}_j$ are the specifications concerning any set of objective functions with or without goals and priorities. Let these ranks be denoted by $rank(\boldsymbol{F}_x, \boldsymbol{S}_i)$ and $rank(\boldsymbol{F}_x, \boldsymbol{S}_j)$, respectively. The “OR” and “AND” operations for the two goal settings are then defined as,

$$rank(\boldsymbol{F}_x, \mathbf{S}_i \cup \mathbf{S}_j) = \min\{rank(\boldsymbol{F}_x, \mathbf{S}_i), rank(\boldsymbol{F}_x, \mathbf{S}_j)\} \quad (7a)$$

$$rank(\boldsymbol{F}_x, \mathbf{S}_i \cap \mathbf{S}_j) = \max\{rank(\boldsymbol{F}_x, \mathbf{S}_i), rank(\boldsymbol{F}_x, \mathbf{S}_j)\} \quad (7b)$$

According to Eq. 7, the rank value of vector $\boldsymbol{F}_x$ for an "OR" operation between any two specifications $\boldsymbol{S}_i$ and $\boldsymbol{S}_j$ takes the minimum rank value with respect to the two specification settings. This is in order to evolve the population towards one of the specifications in which the objective vector is less strongly violated. In contrast, an "AND" operation takes the maximum rank value in order to direct the evolutionary search towards minimizing the amount of violation from both of the specifications concurrently. Clearly, the "AND" and "OR" operations in Eq. 7 can be easily extended to include general logical specifications with more complex connectives, such as "($\boldsymbol{S}_i$ OR $\boldsymbol{S}_j$) AND ($\boldsymbol{S}_k$ OR $\boldsymbol{S}_l$)", if desired.

## 3. Dynamic Sharing Scheme and MOEA Program Flowchart

### 3.1 Dynamic Sharing Scheme

Fitness sharing was proposed by Goldberg and Richardson (1987) to evolve an equally distributed population along the Pareto-optimal front or to distribute the population at multiple optima in the search space. The method creates sub-divisions in the objective domain by degrading an individual fitness upon the existence of other individuals in its neighborhood defined by a sharing distance. The niche count, $m_i = \sum_j^N sh(d_{i,j})$, is calculated by summing a sharing function over all members of the population, where the distance $d_{i,j}$ represents the distance between individual $i$ and $j$. The sharing function is defined as

$$sh(d_{i,j}) = \begin{cases} 1 - \left(\dfrac{d_{i,j}}{\sigma_{share}}\right)^{\alpha} & if \ d_{i,j} < \sigma_{share} \\ 0 & otherwise \end{cases} \quad (8)$$

with the parameter $\alpha$ being commonly set to 1.

The sharing function in Eq. 8 requires a good setting of sharing distance $\sigma_{share}$ to be estimated upon the trade-off surface, which is usually unknown in many optimization problems (Coello Coello, 1999). Moreover, the size of objective space usually cannot be predefined, as the exact bounds of the objective space are often undetermined. Fonseca and Fleming (1993) proposed the method of Kernel density estimation to determine an appropriate sharing distance for MO optimization. However, the sharing process is performed in the 'sphere' space which may not reflect the actual objective space for which the population is expected to be uniformly distributed. Miller and Shaw (1996) proposed a dynamic sharing method for which the peaks in the parameter domain are 'dynamically' detected and recalculated at every generation with the sharing distance remains predefined. However, the approach is made on the assumption that the number of niche peaks can be estimated and the peaks are all at the minimum distance of $2\sigma_{share}$ from each other.

Moreover, their formulation is defined in the parameter space to handle multi-modal function optimization, which may not be appropriate for distributing the population uniformly along the Pareto-optimal front in the objective domain.

In contrast to existing approaches, we propose a dynamic sharing method that adaptively computes the sharing distance $\sigma_{share}$ to uniformly distribute all individuals along the Pareto-optimal front at each generation. This requires no prior knowledge of the trade-off surface. Intuitively, the trade-offs for an $m$-objective optimization problem are in the form of an ($m$-1) dimensional hyper-volume (Tan et al. 1999), which can be approximated by the hyper-volume $V_{pop}{}^{(n)}$ of a hyper-sphere as given by,

$$V_{pop}{}^{(n)} = \frac{\pi^{(m-1)/2}}{\left(\frac{m-1}{2}\right)!} \times \left[\frac{d^{(n)}}{2}\right]^{m-1} \tag{9}$$

where $d^{(n)}$ is the diameter of the hyper-sphere at generation $n$. Note that computation of the diameter $d^{(n)}$ depends on the curvature of the trade-off curve formed by the non-dominated individuals in the objective space. For a two-objective optimization problem, the diameter $d^{(n)}$ is equal to the interpolated distance of the trade-off curve covered by the non-dominated individuals as shown in Figure 4. Although computation of $d^{(n)}$ that accurately represents the interpolated curvature of the non-dominated individuals distribution is complex, it can be estimated by the average distance between the shortest and the longest possible diameter given by $d_{min}{}^{(n)}$ and $d_{max}{}^{(n)}$ respectively (Tan et al. 1999). Let $\boldsymbol{F}_x$ and $\boldsymbol{F}_y$ denote the objective function of the two furthest individuals in a population. Then $d_{min}{}^{(n)}$ is equal to the minimum length between $\boldsymbol{F}_x$ and $\boldsymbol{F}_y$, and $d_{max}{}^{(n)}$ can be estimated by $d_1{}^{(n)} + d_2{}^{(n)}$ as shown in Figure 4.

The same procedure can also be extended to any multi-dimensional objective space. To achieve a uniformly distributed population along the trade-off set, the sharing distance $\sigma_{share}{}^{(n)}$ could be computed as half of the distance between each individual in the ($m$-1)-dimensional hyper-volume $V_{pop}{}^{(n)}$ covered by the population size $N$ at generation $n$,

$$N \times \frac{\pi^{(m-1)/2}}{\left(\frac{m-1}{2}\right)!} \times \left(\sigma_{share}^{(n)}\right)^{m-1} = V_{pop}^{(n)} \tag{10}$$

Substituting Eq. 9 into Eq. 10 gives the sharing distance $\sigma_{share}{}^{(n)}$ at generation $n$ in term of the diameter $d^{(n)}$ and the population size $N$ as given by

$$\sigma_{share}^{(n)} = N^{1/(1-m)} \times \frac{d^{(n)}}{2} \tag{11}$$

Clearly, Eq. 11 provides a simple computation of $\sigma_{share}$ that is capable of distributing the population evenly along the Pareto front, without the need for any prior knowledge of the usually unknown fitness landscape. Moreover, adopting the computation of sharing distance that is dynamically based upon the population distribution at each generation is also more appropriate and effective than the method of off-line estimation with pre-assumed trade-off surface as employed in most existing sharing methods, since the trade-off surface may be changed any time along the evolution whenever the goal setting is altered.

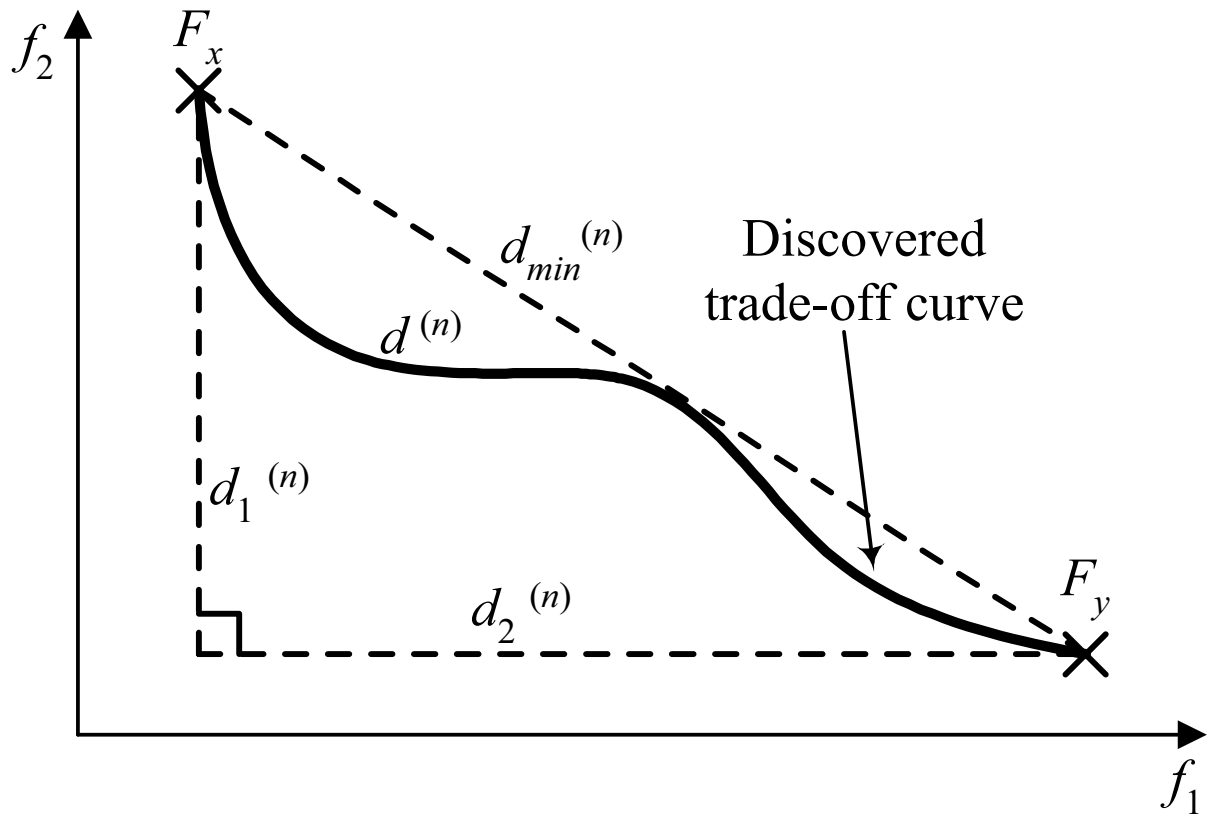

Figure 4: The Diameter $d^{(n)}$ of a Trade-off Curve

### 3.2 MOEA Program Flowchart

The overall program flowchart of this paper's multi-objective evolutionary algorithm (MOEA) is illustrated in Figure 5. At the beginning of the evolution, a population of candidate solutions is initialized and evaluated according to a vector of objective functions. Based upon the user-defined specifications, such as goals, constraints, priorities and logical operations, the evaluated individuals are ranked according to the goal-sequence domination scheme (described in Section 2) in order to evolve the search towards the global trade-off surface. The resulted rank values are then further refined by the dynamic sharing scheme (described in Section 3.1) in order to distribute the non-dominated individuals uniformly along the discovered Pareto-optimal front. If the stopping criterion is not met, the individuals will undergo a series of genetic operations which are detailed within the "genetic operations" in Figure 6. Here, simple genetic operations consisting of tournament selection (Tan et al. 1999), simple crossover with mating restriction that selects individuals within the sharing distance for mating (Fonseca & Fleming, 1998) as well as simple mutation are performed to reproduce offspring for the next generation.

After the genetic operations, the newly evolved population is evaluated and combined with the non-dominated individuals preserved from the previous generation. The combined population is then subjected to the domination comparison scheme and pruned to the desired population size according to the Switching Preserved Strategy (SPS) (Tan et al. 1999). This maintains a set of stable and well-distributed non-dominated individuals along the Pareto-optimal front. In SPS, if the number of non-dominated individuals in the combined population is less than or equal to the desired population size, extra individuals are removed according to their rank values in order to promote stability in the evolutionary search towards the final trade-offs. Otherwise, the non-dominated individuals with high niched count value will be discarded in order to distribute the individuals uniformly along the discovered Pareto-optimal front. After the process, the remaining individuals are allowed to survive in the next generation and this evolutionary cycle is repeated until the stopping criterion is met.

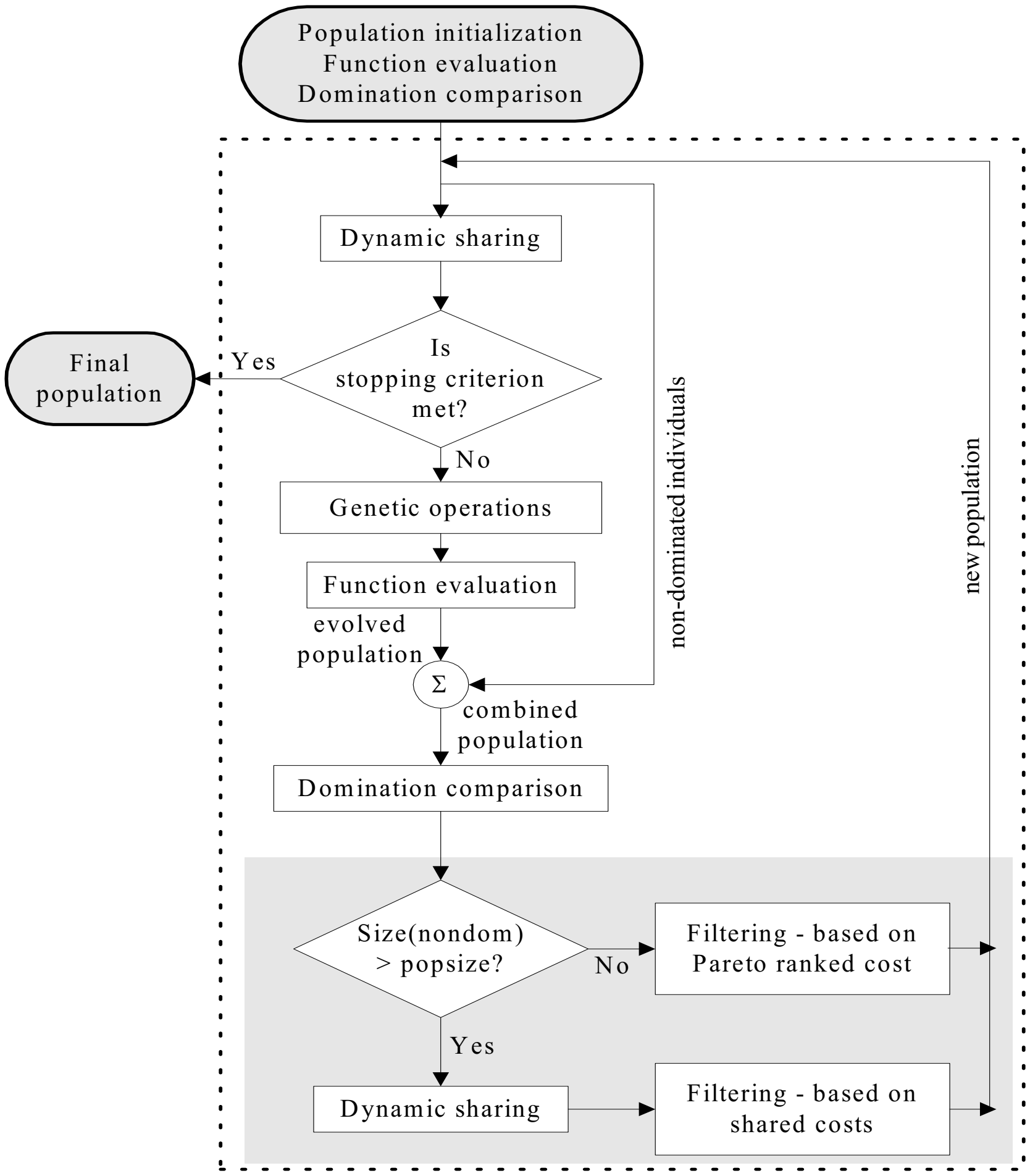

Figure 5: Program Architecture of the MOEA

Genetic Operations for MOEA:

Let, $pop^{(n)}$ = population in current generation $n$

Step 1) Perform tournament selection to select individuals from $pop^{(n)}$. The selected population is called $selpop^{(n)}$.

Step 2) Perform simple crossover and mating restriction for $selpop^{(n)}$ using the dynamic sharing distance in Step 1. The resulted population is called $crosspop^{(n)}$.

Step 3) Perform simple mutation for $crosspop^{(n)}$. The resulted population is called $evolpop^{(n)}$.

Figure 6: Detailed procedure within the box of “genetic operations” in Figure 5

## 4. Validation Results on Benchmark Problems

This section validates the proposed algorithm in two ways. The first kind of validation (presented in Section 4.1) illustrates how each of the proposed features, including goal-sequence domination scheme, hard/soft goal and priority specifications, logical operations among multiple goals and dynamic sharing, enhances the performance of MOEA in MO optimization. As shown in Section 4.2, the second type of validation compares performance of the proposed MOEA with various evolutionary algorithms based upon a benchmark problem. Various performance measures are used in the comparison and the results are then discussed.

### 4.1 Validation of the Proposed Features in MOEA

In this section, various proposed features in MOEA are examined for their usefulness in MO optimization. This study adopts the simple two-objective minimization problem (Fonseca & Fleming, 1993) that allows easy visual observation of the optimization performance. The function has a large and non-linear trade-off curve, which challenges the algorithm's ability to find and maintain the entire Pareto-optimal front uniformly. The two-objective functions, $f_1$ and $f_2$, to be minimized are given as

$$\begin{aligned} f_1(x_1,...,x_8) &= 1 - exp\left(-\sum_{i=1}^{8}\left(x_i - \frac{1}{\sqrt{8}}\right)^2\right) \\ f_2(x_1,...,x_8) &= 1 - exp\left(-\sum_{i=1}^{8}\left(x_i + \frac{1}{\sqrt{8}}\right)^2\right) \end{aligned} \tag{12}$$

where, $-2 \le x_i < 2, \forall i = 1,2,...,8$. The trade-off line is shown by the curve in Figure 7, where the shaded region represents the unfeasible area in the objective domain.

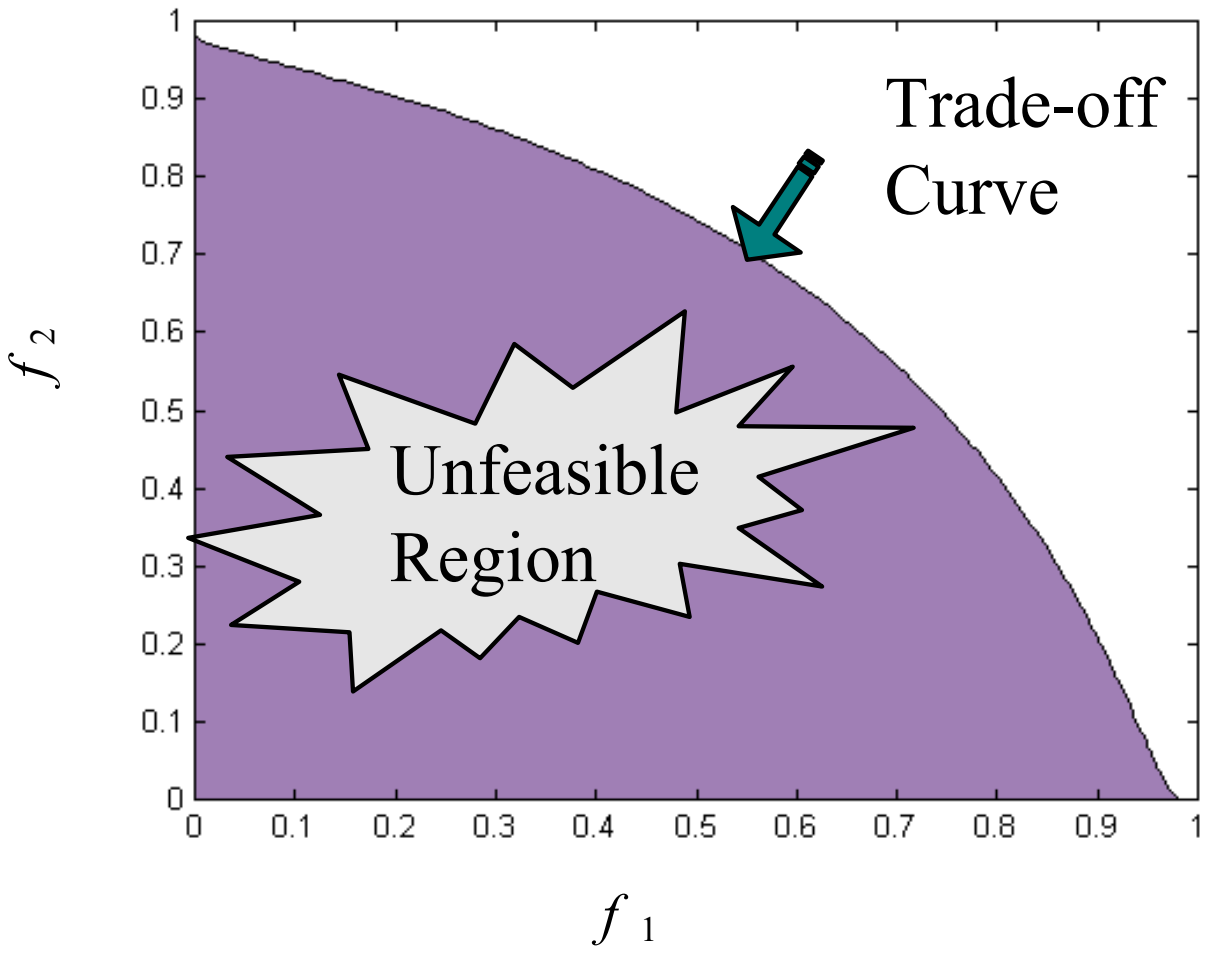

Figure 7: Pareto-optimal Front in the Objective Domain

The simulations are run for 70 generations with a population size of 100. Standard mutation with a probability of 0.01 and standard two-point crossover with a probability of 0.7 are used. To study the merit of the dynamic sharing scheme in MOEA as proposed in Section 3.1, 4 different types of simulations have been performed. The first type is without fitness sharing. The second and third employ a fixed sharing distance of 0.01 and 0.1, respectively. The fourth uses the dynamic sharing scheme which does not require any predefined sharing distance setting. Figure 8 illustrates the respective population distribution in the objective domain at the end of the evolution. It can be observed that all of the four simulations are able to discover the final trade-off, but with some performance difference in terms of the closeness and uniformity of the population distribution along the trade-off curve.

For the MOEA without fitness sharing as shown in Figure 8a, the population tends to converge to an arbitrary part of the trade-off curve. This agrees with the findings of Fonseca and Fleming, (1993). For the MOEA with fitness sharing, as shown in Figures 8b and 8c, the population can be distributed along the trade-off curve rather well, although the sharing distance of 0.01 provides a more uniform distribution than that of 0.1. This indicates that although fitness sharing contributes to population diversity and distribution along the trade-off curve, the sharing distance has to be chosen carefully in order to ensure the uniformity of the population distribution. This often involves tedious trial-and-error procedures in order to 'guess' an appropriate sharing distance, since it is problem dependent and based upon the size of the discovered trade-offs as well as the number of non-dominated individuals. These difficulties can be solved with the proposed dynamic sharing scheme, which has the ability to automatically adapt the sharing distance along the evolution without the need of any predefined parameter, as shown in Figure 8d.

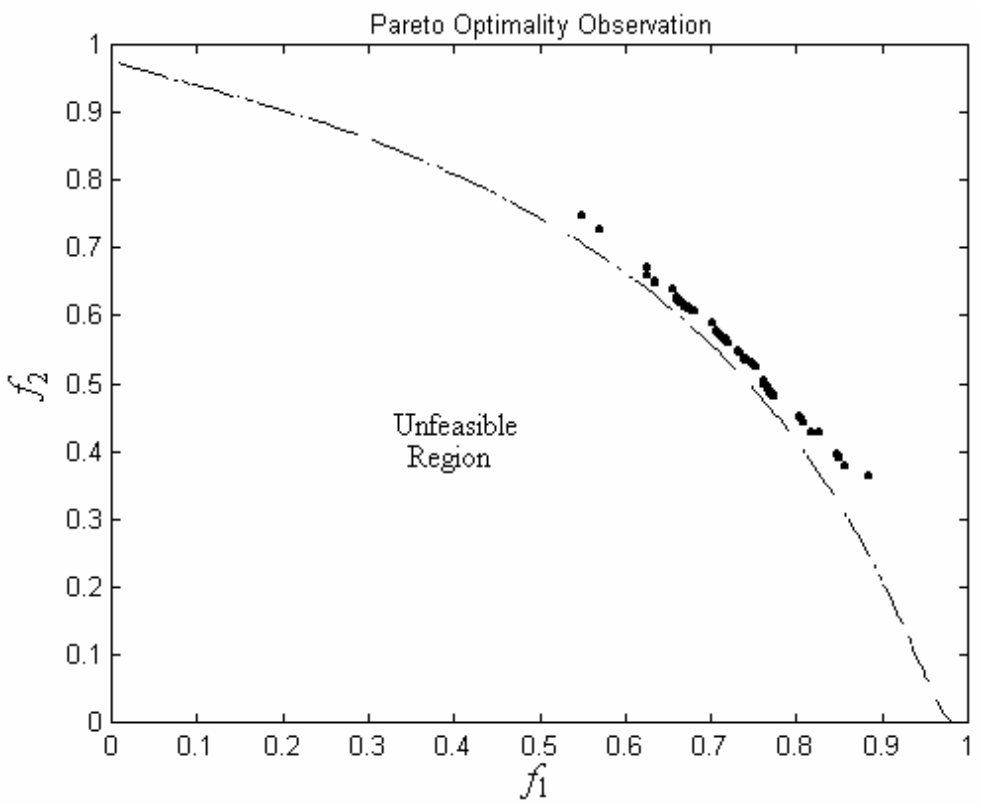

(a) No sharing

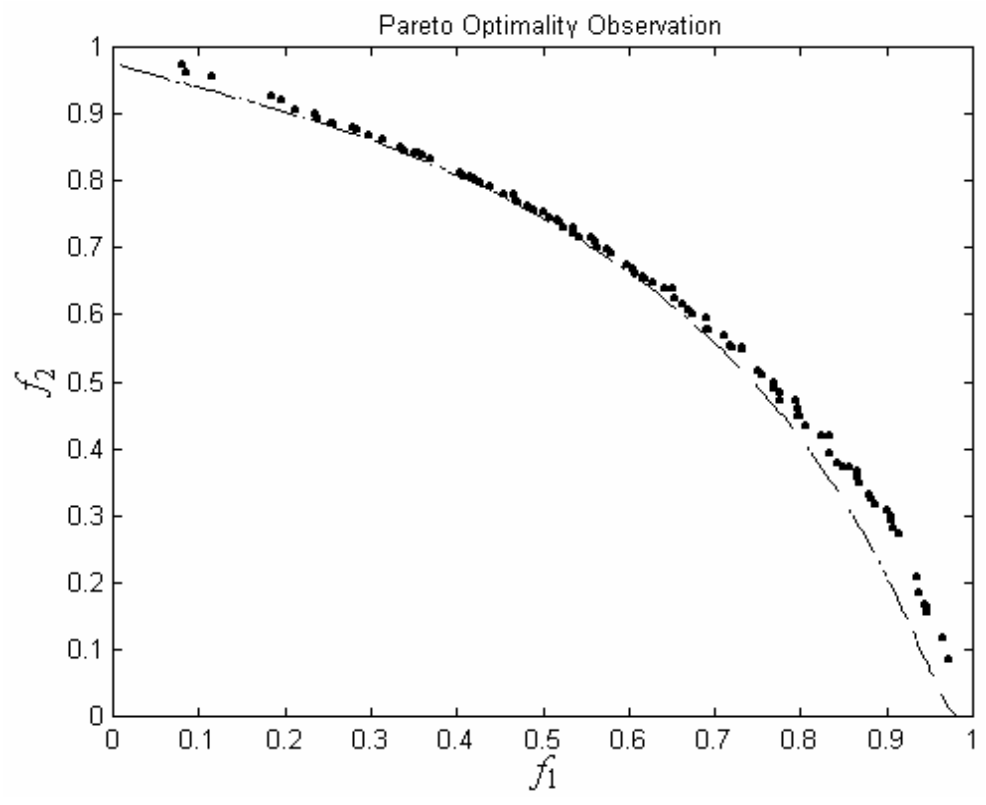

(b) Sharing distance = 0.01

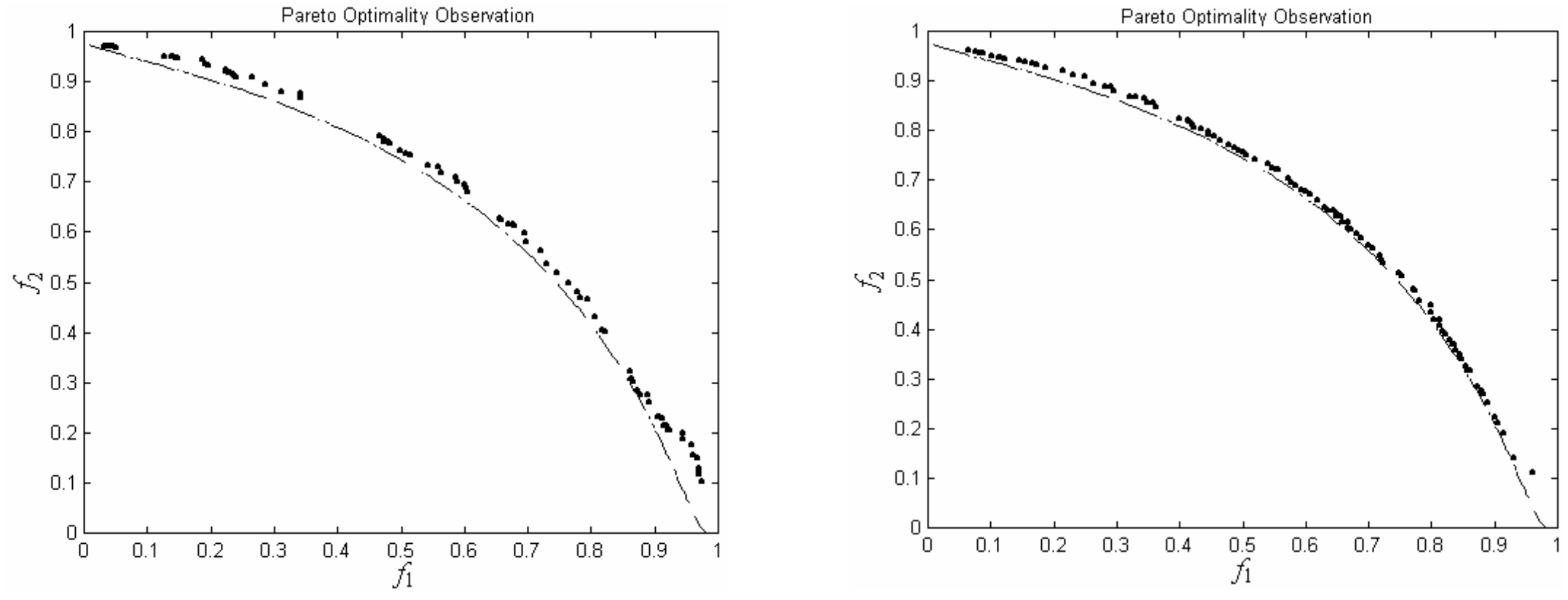

(c) Sharing distance = 0.1 (d) Dynamic sharing

Figure 8: Performance Validation of Dynamic Sharing Scheme in MOEA

To validate the contribution of the switching preserved strategy (SPS) in MOEA, the above simulation was repeated with different scenarios and settings. Figure 9a depicts the simulation result without the implementation of SPS, in which the evolution faces difficulty converging to the trade-off curve. The solid dots represent the non-dominated individuals while the empty circles represent the dominated individuals. As can be seen, the final population is crowded and the non-dominated individuals are distributed with some distance away from the trade-off curve. Figure 9b shows the simulation result for the MOEA with SPS and filtering solely based upon the Pareto domination. The final population has now managed to converge to the Pareto-optimal front. However, the non-dominated individuals are not equally distributed and the diversity of the population is poor: they only concentrate on a portion of the entire trade-off curve (c.f. Figures 8d, 9b). These results clearly show that SPS in MOEA is necessary in order to achieve good stability and diversity of the population in converging towards the complete set of trade-offs.

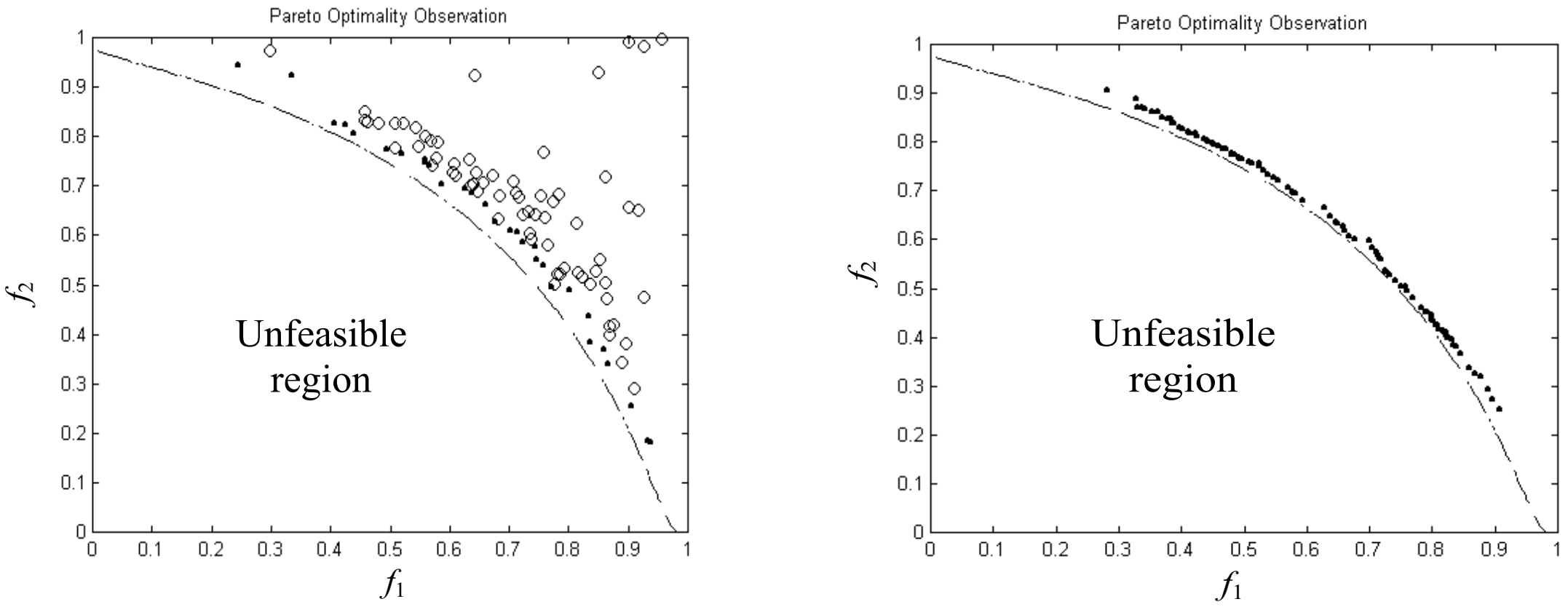

(a) Without SPS (b) With SPS solely based on Pareto ranked cost

Figure 9: Performance Validation of SPS in MOEA

The proposed goal-sequence domination scheme was also validated for problems with different goal settings, including a feasible but extreme goal setting of (0.98, 0.2) and an unfeasible goal setting of (0.7, 0.4) as shown in Figures 10 and 11, respectively. As desired, the population is seen to concentrate on the preferred region of the trade-off curve at the end of the evolution, regardless of the unattainable or extreme goal settings. As shown in Figures 10 and 11, MOEA is capable of uniformly distributing the non-dominated individuals along any trade-offs size resulting from different goal settings, with the help of the dynamic sharing scheme that automatically computes a suitable sharing distance for optimal population distribution at each generation.

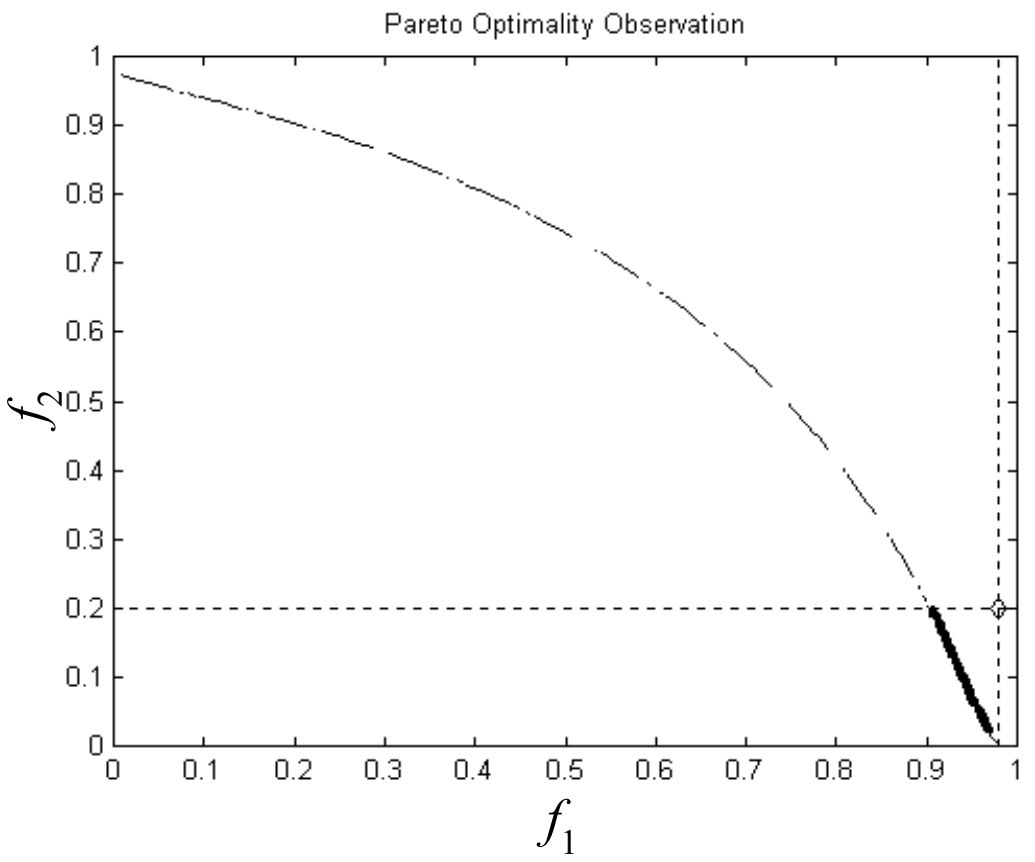

Figure 10: Feasible but Extreme Goal Setting

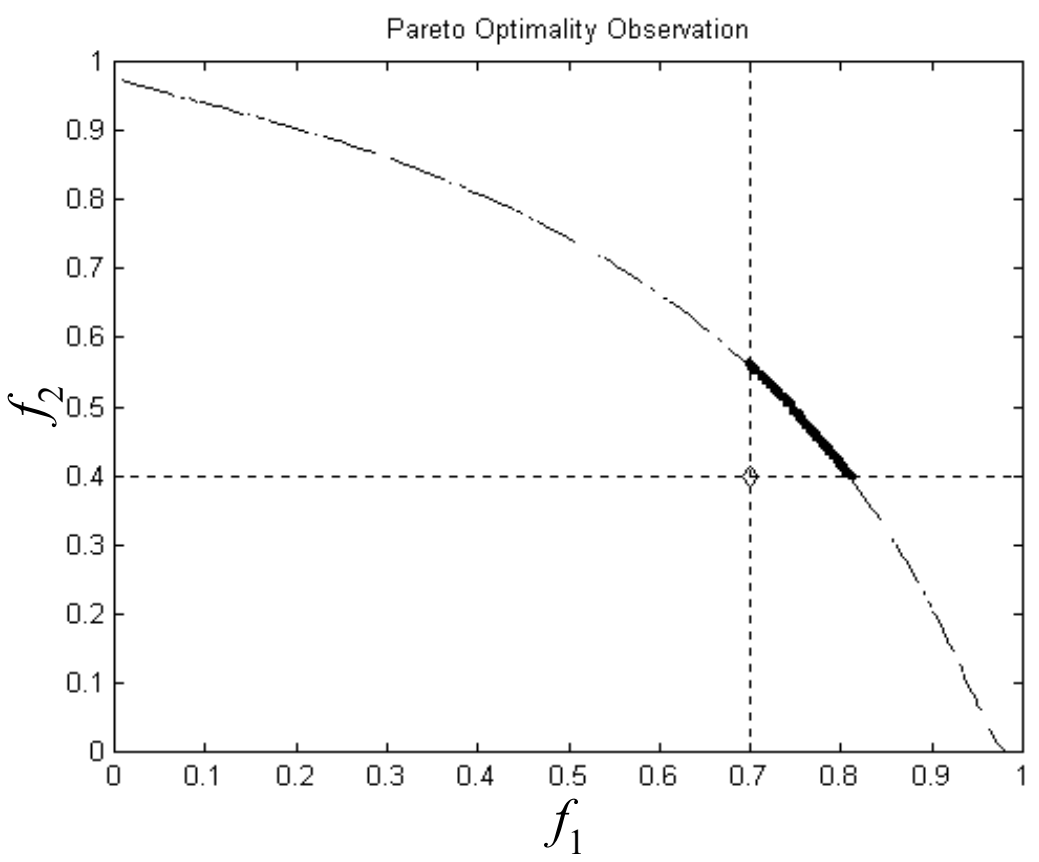

Figure 11: Unfeasible Goal Setting

Figure 12 shows the trace of sharing distance during the evolution. The thin and thick lines represent the average sharing distance without any goal setting (see Figure 8d for the corresponding Pareto-front) and with the goal setting of (0.7, 0.4) (see Figure 11 for the corresponding Pareto-front), respectively. Generally, MO optimization without a goal setting has an initially small size of discovered Pareto-front, which subsequently grows along with the evolution to approach and cover the entire trade-off region at the end of evolution. This behavior is explained in Figure 12 where the sharing distance increases asymptotically along the evolution until a steady value of 0.0138 is reached. It should be noted that this value is close to the fixed sharing distance of 0.01 in Figure 8b, which was carefully chosen after trial-and-error procedures. For the case of MOEA with a goal setting of (0.7, 0.4), the sharing distance increases initially and subsequently decreases to 0.0025 along the evolution, which is lower than the value of 0.0138 (without goal setting). The reason is that the concentrated trade-off region within the goal setting is smaller than the entire trade-off region (without goal setting), and hence results in a smaller distance for uniform sharing of non-dominated individuals. These experiments show that the proposed dynamic sharing scheme can effectively auto-adapt the sharing distance to arrive at an appropriate value for uniform population distribution along the discovered trade-off region at different sizes, without the need for any a priori parameter setting.

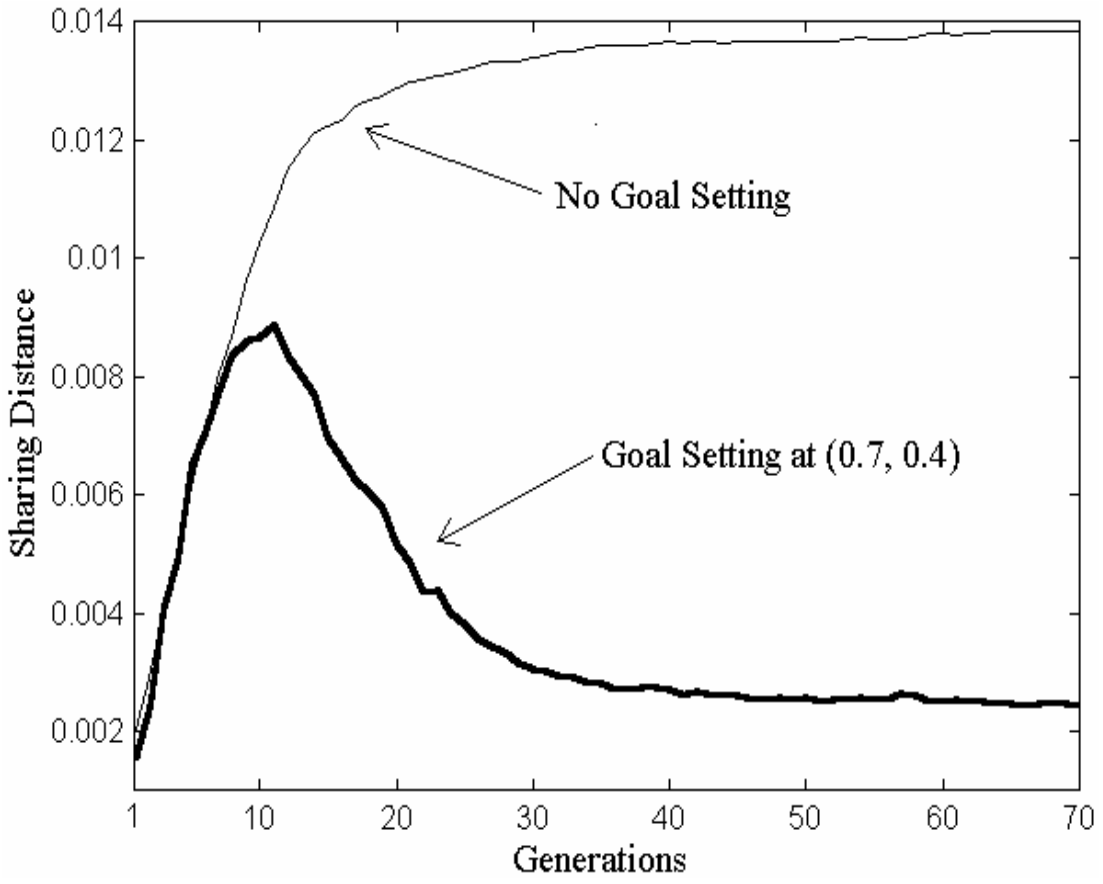

Figure 12: Trace of the Dynamic Sharing Distance Along the Evolution

Figures 13 and 14 show the MOEA simulation results for the case of an infeasible goal setting with soft and hard priorities, respectively. In the figures, diamonds represent goals, small circles represent non-dominated individuals and solid dots represent dominated individuals. For the soft priority setting in Figure 13, goals are treated as first priority followed by the objective component of $f_1$ as second priority, i.e., $\boldsymbol{P}_g = [1, 1]$ and $\boldsymbol{P}_f = [2, 0]$. As can be seen, it provides a distributive optimization approach for all goals by pushing the population towards the objective component of $f_1$ that has a higher priority, after taking the goal vector into consideration (c.f. Figures 3a, 13b). In contrast, Figure 14 shows the minimization results with hard priority setting where priority of $f_1$ is higher than $f_2$, i.e., $\boldsymbol{P}_g = [1, 2]$ and $\boldsymbol{P}_f = [0, 0]$. Unlike the soft priority optimization, hard priority minimizes the objective of $f_1$ until the relevant goal component of $g_1 = 0.5$ is satisfied before attaining the objective component of $f_2$ with the second goal component of $g_2 = 0.5$, as shown in Figure 14 (c.f. Figures 3b, 14b). As can be seen, objective values with hard priority settings are better with higher priority but are worse with lower priority, as compared to the solutions obtained in soft priority optimization (c.f. Figures 13b, 14b).

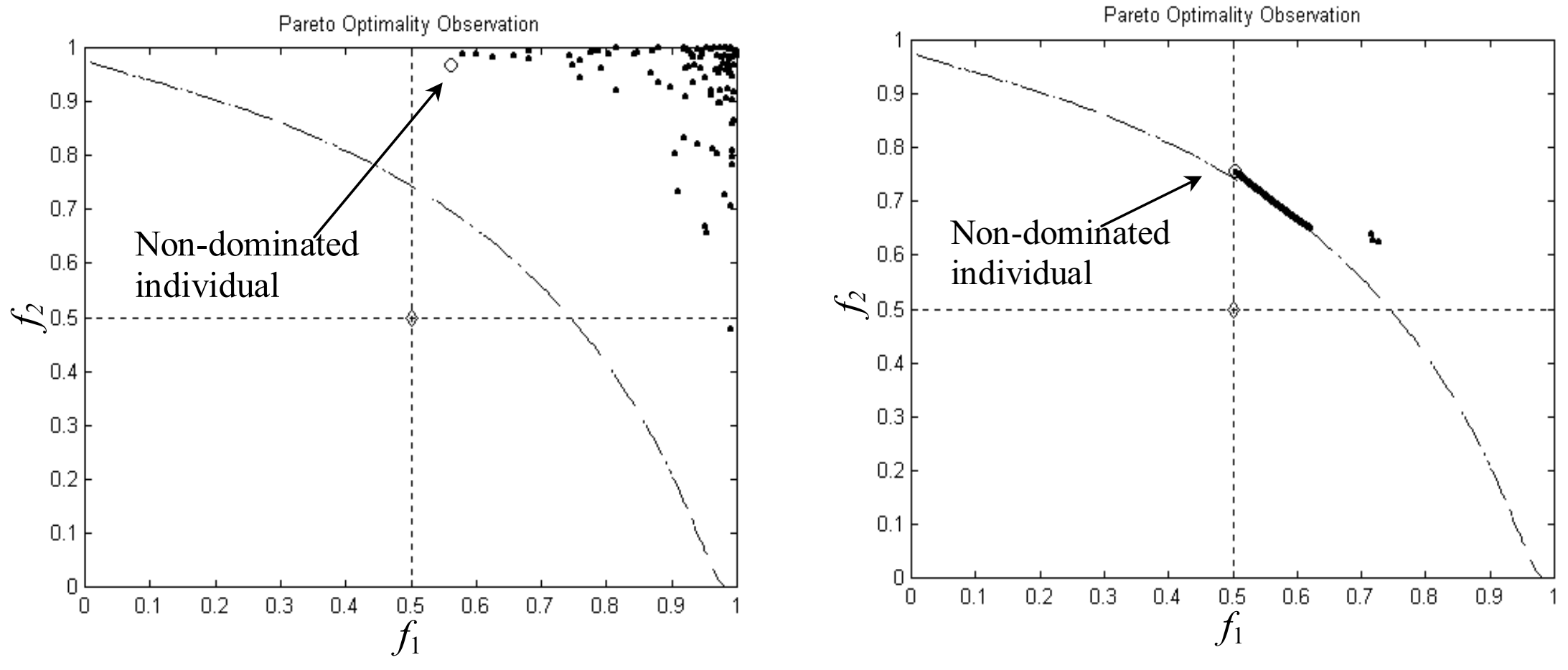

(a) At generation 5 (b) At generation 70

Figure 13: MOEA Optimization with Unfeasible Goal Setting: $f_1$ has Soft Priority Higher than $f_2$

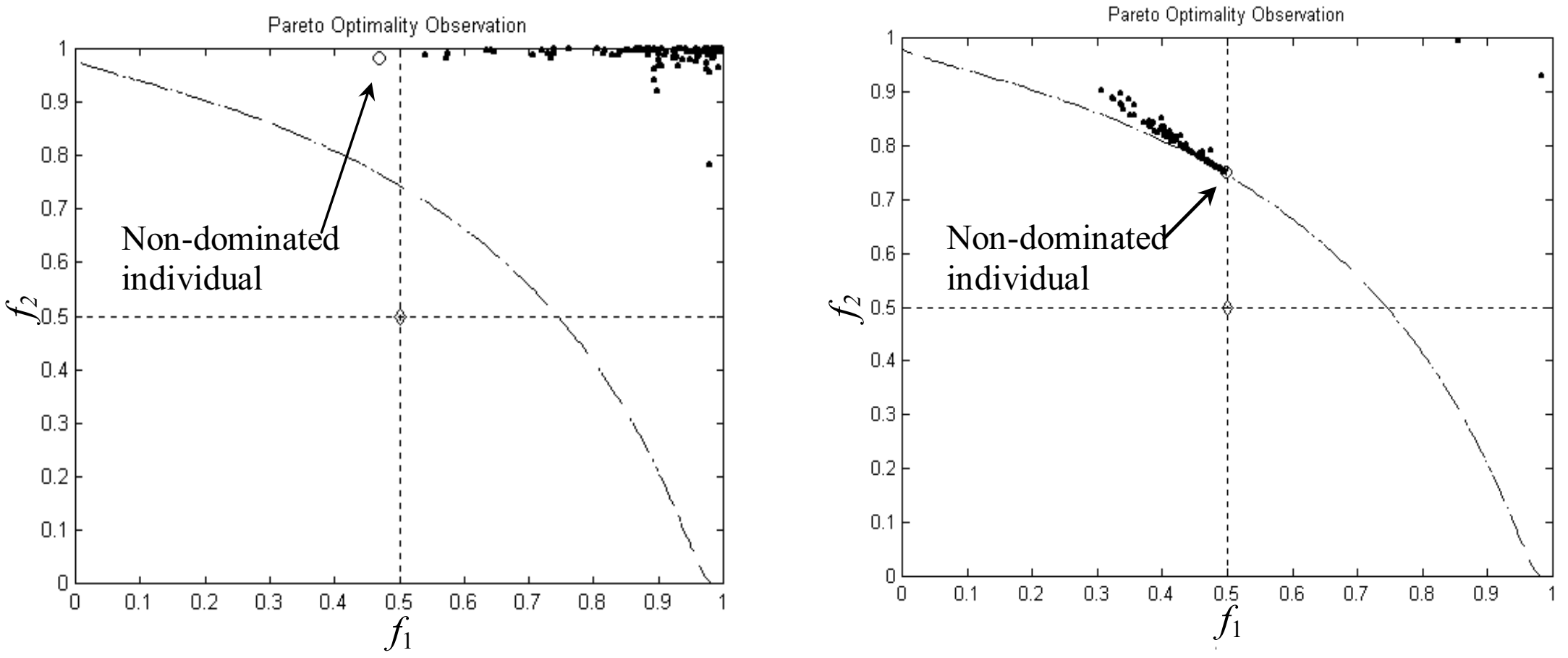

(a) At generation 5 (b) At generation 70

Figure 14: MOEA Optimization with Unfeasible Goal Setting: $f_1$ has Hard Priority Higher than $f_2$

Figure 15 shows the MOEA minimization result with $f_1$ being a hard constraint. The population continuously evolves towards minimizing $f_2$ only after the hard constraint of $f_1$ has been satisfied. In general, objective components with hard constraints may be assigned as hard priorities in order to meet the hard constraints before minimizing any other objective components.

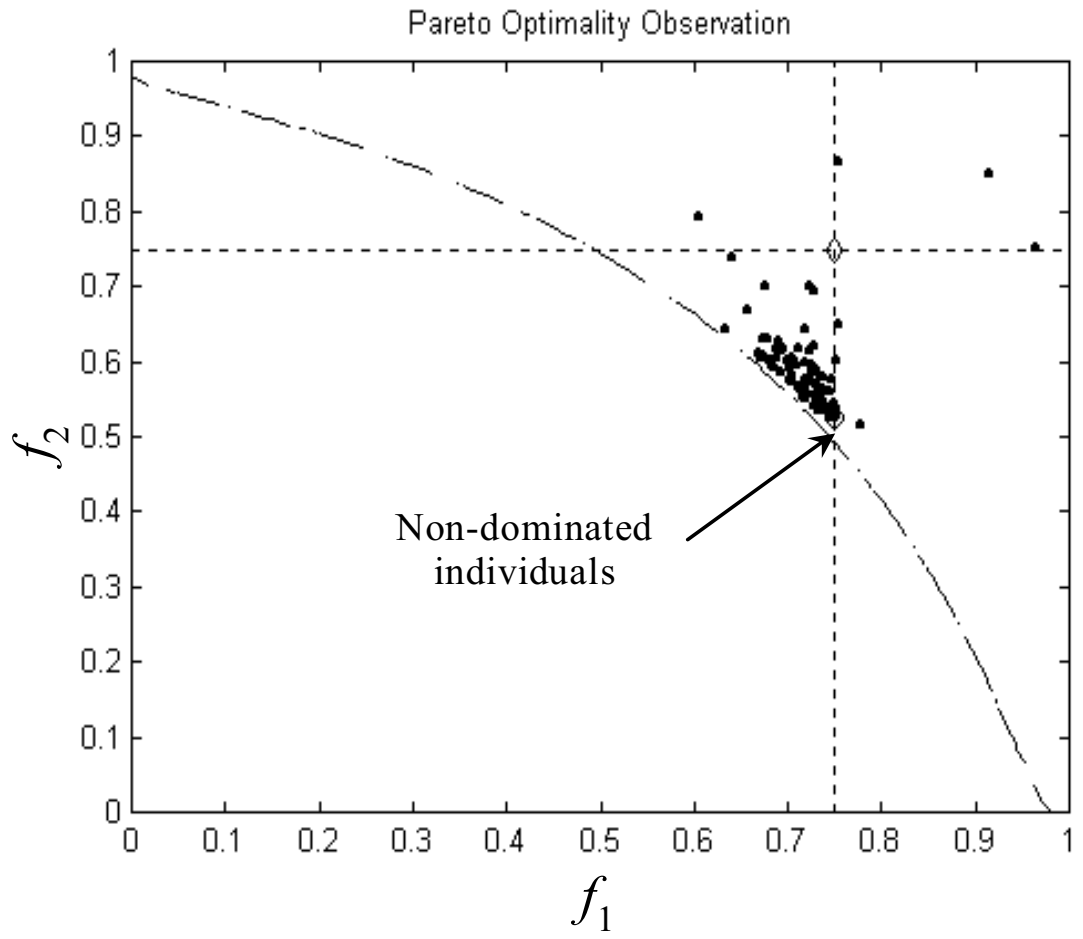

Figure 15: MOEA Minimization with Hard Constraint on $f_1$

Figures 16 and 17 show the MO optimization results that include multiple goal settings specified by logical "OR" ( $\cup$ ) and "AND" ( $\cap$ ) connectives, respectively. For the "OR" operation as shown in Figure 16, the population is automatically distributed and equally spread over the different concentrated trade-off regions to satisfy the goal settings separately, regardless of the overlapping or feasibility of the goals. With the proposed dynamic sharing scheme, the sub-population size for each goal is in general based upon the relative size of the concentrated trade-off surface of that goal, and thus individuals are capable of equally distributing themselves along the different concentrated trade-off regions. For the “AND” operation as illustrated in Figure 17, the whole population evolves towards minimizing all the goals $\boldsymbol{G}_1$, $\boldsymbol{G}_2$ and $\boldsymbol{G}_3$ simultaneously. As a result, the individuals are equally distributed over the common concentrated trade-off surface formed by the three goals, as desired.

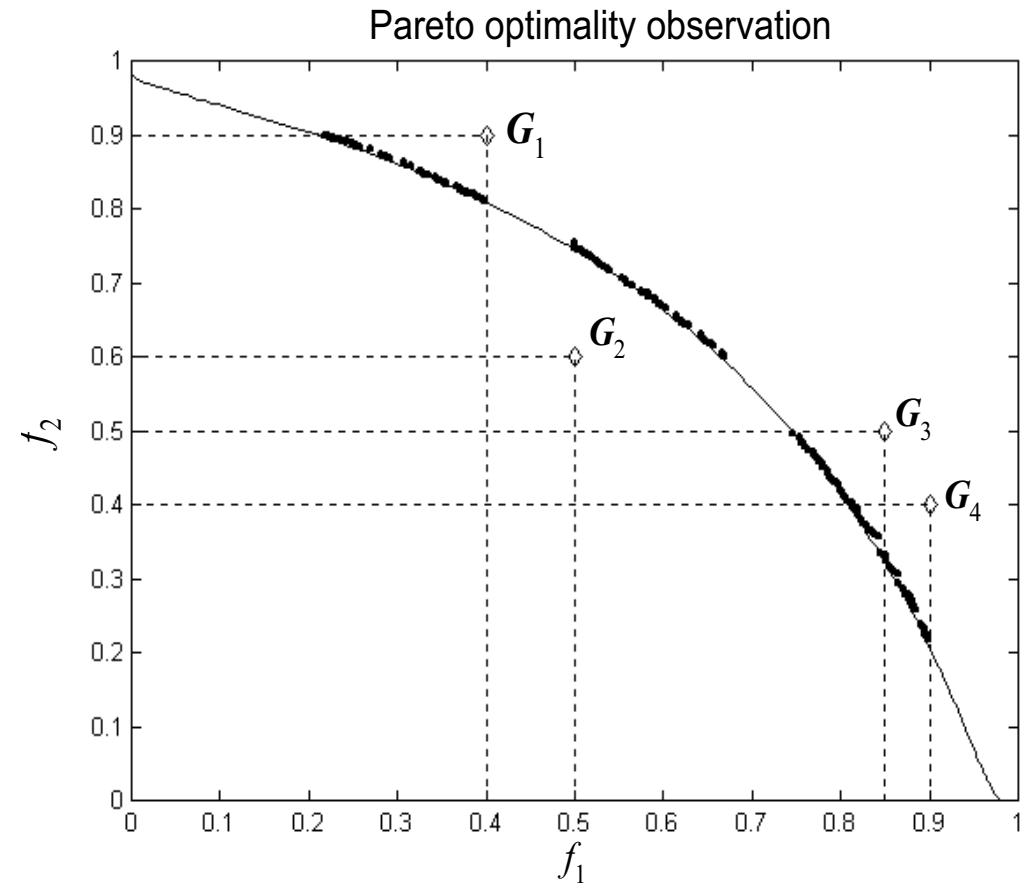

Figure 16: MOEA Minimization for ($\boldsymbol{G}_1 \cup \boldsymbol{G}_2 \cup \boldsymbol{G}_3 \cup \boldsymbol{G}_4$)

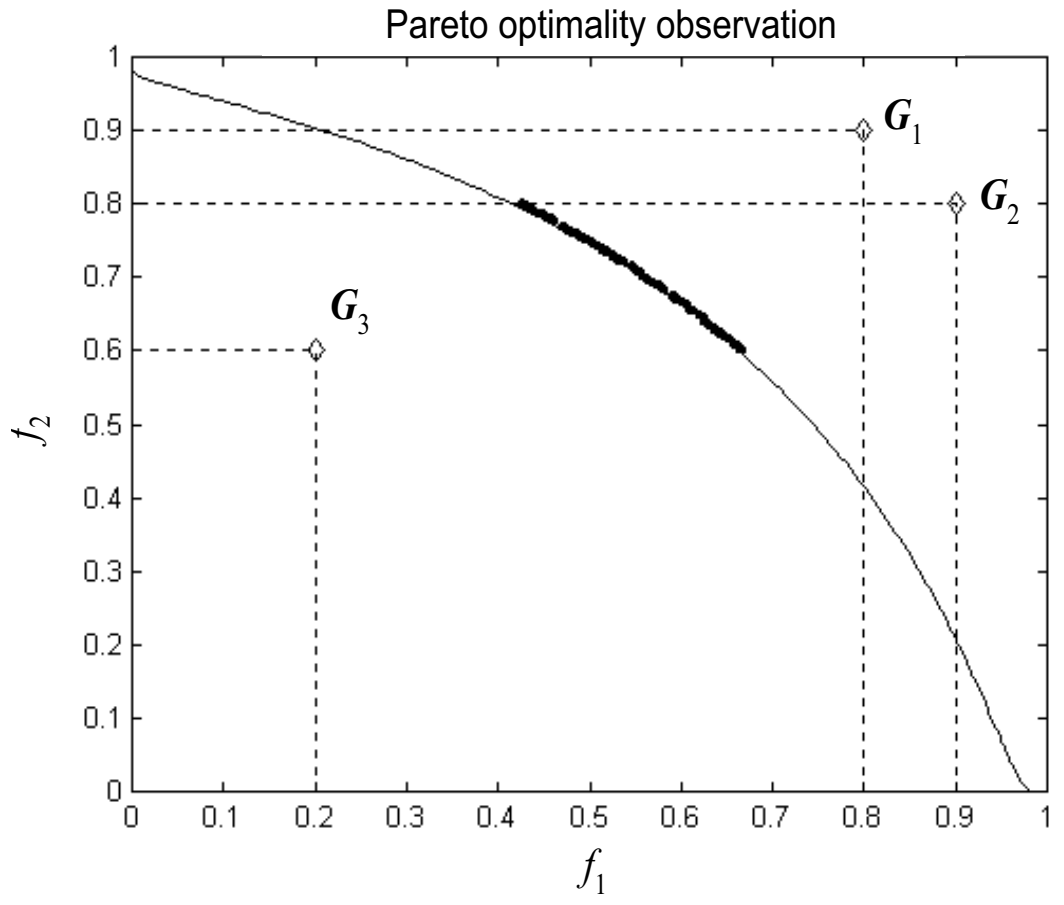

Figure 17: MOEA Minimization for ($\boldsymbol{G}_1 \cap \boldsymbol{G}_2 \cap \boldsymbol{G}_3$)

### 4.2 Performance Comparisons of MOEA

This section studies and compares the performance of the proposed MOEA with other multi-objective evolutionary optimization methods based upon a benchmark MO optimization problem. For a comprehensive comparison, various performance measures are used and the comparison results are discussed in the section.

#### 4.2.1 The Test Problem

The test problem used for the performance comparisons is a two-objective minimization problem (Deb, 1999). The problem is chosen because it has a discontinuous Pareto-front which challenges the evolutionary algorithm's ability to find and maintain the Pareto-optimal solutions that are discontinuously spread in the search space. The problem involves minimizing the objective functions $f_1$ and $f_2$ as given below,

$$f_1(x_1) = x_1 \tag{13a}$$

$$g(x_2,\ldots,x_{10}) = 1 + 10\frac{\sum_{i=2}^{10} x_i}{10-1}, \tag{13b}$$

$$h(f_1, g) = 1 - (f_1/g)^{0.25} - (f_1/g)\sin(10\pi f_1) \tag{13c}$$

$$f_2(x_1) = g(x_2,\ldots,x_{10})h(f_1, g) \tag{13d}$$

All variables are varied in [0, 1] and the true Pareto-optimal solutions are constituted with $x_i = 0$ $\forall\ i = 2, \ldots, 10$ and the discontinuous values of $x_1$ in the range of [0, 1] (Deb, 1999). Figure 18 depicts the discontinuous Pareto-optimal front (in bold). The shaded region represents the unfeasible region in the search space.

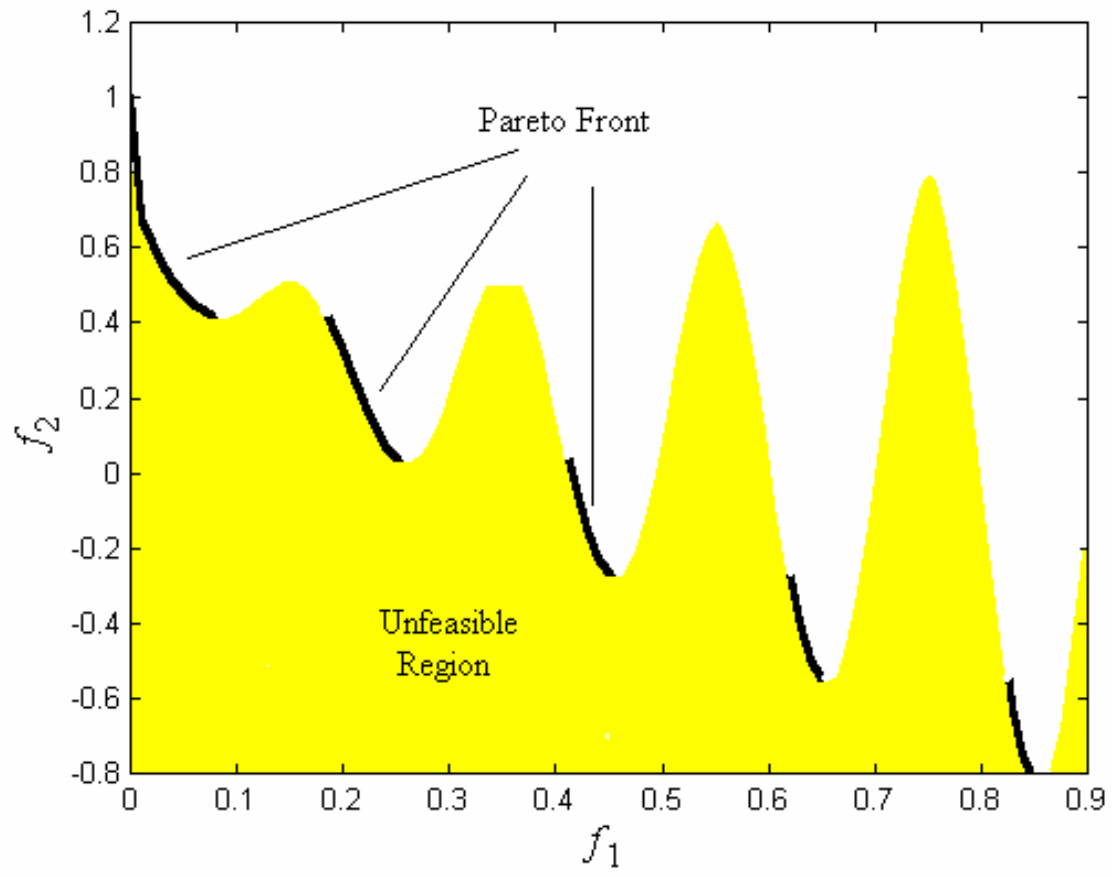

Figure 18: Pareto-optimal Front in the Objective Domain

### 4.2.2 Current Evolutionary MO Optimization Methods

Besides MOEA, five well-known multi-objective evolutionary optimization methods are used in the comparison. These approaches differ from each other in their working principles and mechanisms and have been widely cited or applied to real-world applications. The algorithms are summarized below and readers may refer to their respective references for detailed information.

(i) *Fonseca and Fleming's Genetic Algorithm* (*FFGA*): For MO optimization, Fonseca and Fleming (1993) proposed a multi-objective genetic algorithm (MOGA) with Pareto-based ranking scheme, in which the rank of an individual is based on the number of other individuals in the current population that dominate it. Their algorithm was further incorporated with fitness sharing and mating restriction to distribute the population uniformly along the Pareto-optimal front.

(ii) *Niched Pareto Genetic Algorithm* (*NPGA*): The method of NPGA (Horn & Nafpliotis, 1993) works on a Pareto-dominance-based tournament selection scheme to handle multiple objectives simultaneously. To reduce the computational effort, a pre-specified number of individuals are picked as a comparison set to help determine the dominance. When both competitors end in a tie, the winner is decided through fitness sharing (Goldberg and Richardson, 1987).

(iii) *Strength Pareto Evolutionary Algorithm* (*SPEA*): The main features of SPEA (Zitzler & Thiele, 1999) are the usage of two populations (*P* and *P'*) and clustering. In general, any non-dominated individual is archived in *P'* and any dominated individual that is dominated by other members in *P'* is removed. When the number of individuals in *P'* exceeds a maximum value, clustering is adopted to remove the extra individuals in *P'*. Tournament selection is then applied to reproduce individuals from *P* + *P'* before the evolution proceeds to the next generation.

(iv) *Non-Generational Genetic Algorithm* (*NGGA*): In NGGA (Borges & Barbosa, 2000), a cost function of an individual is a non-linear function of domination measure and density measure on that individual. Instead of evolving the whole population at each iteration, a pair of parents is selected to reproduce two offsprings. An offspring will replace the worst individual in a population if the offspring has lower cost function than the worst individual.

(v) *Murata and Ishibuchi's Genetic Algorithm* (*MIGA*): Unlike the above evolutionary optimization methods, MIGA (Murata & Ishibuchi, 1996) applies the method of weighted-sum to construct the fitness of each individual in a population. To keep the diversity of the population along the Pareto-optimal front, the weights are randomly specified when a pair of parent solutions is selected from a current population for generating the offspring.

### 4.2.3 Performance Measures

This section considers three different performance measures which are complementary to each other: Size of space covered (*SSC*), uniform distribution (*UD*) index of non-dominated individuals and number of function evaluation (*Neval*).

(i) *Size of Space Covered* (*SSC*): This measure was proposed by Zitzler and Thiele (1999) as a measure to quantify the overall size of phenotype space covered (*SSC*) by a population. In general, the higher the value of *SSC*, the larger the space covered by the population and hence the better the optimization result.

(ii) *Uniform Distribution* (*UD*) *of Non-dominated Population*: Besides the size of space covered by a population, it is also essential to examine the ability of an evolutionary optimization to distribute their non-dominated individuals as uniformly as possible along the discovered

Pareto-optimal front, unless prohibited by the geometry of the Pareto front. This is to achieve a smooth transition from one Pareto-optimal solution to its neighbors, thus facilitating the decision-maker in choosing his/her final solution. Mathematically, $UD(\mathbf{X}')$ for a given set of non-dominated individuals $\mathbf{X}'$ in a population $\mathbf{X}$, where $\mathbf{X}' \subseteq \mathbf{X}$, is defined as (Tan et al. 2001a),

$$UD(\mathbf{X}') = \frac{1}{1 + S_{nc}} \tag{14}$$

where $S_{nc}$ is the standard deviation of niche count of the overall set of non-dominated individuals $\mathbf{X}'$. It can be seen that larger value of $UD(\mathbf{X}')$ indicates a more uniform distribution and vice versa.

(iii) *Number of Function Evaluation* (*Neval*): The computational effort required to solve an optimization problem is often an important issue, especially when only limited computing resources are available. In the case that a fixed period of CPU time is allocated and the CPU time for each function evaluation is assumed to be equal, then more function evaluations being performed by an optimization indirectly indicates less additional computational effort is required by the algorithm.

### 4.2.4 Simulation Settings and Comparison Results

The decimal coding scheme (Tan et al. 1999) is applied to all the evolutionary methods studied in this comparison, where each parameter is coded in 3-digit decimals and all parameters are concatenated together to form a chromosome. In all cases, two-point crossover with a probability of 0.07 and standard mutation with a probability of 0.01 are used. A reproduction scheme is applied according to the method used in the original literature of each algorithm under comparison. The population size of 100 is used in FFGA, NPGA, NGGA and MOEA, which only require a single population in the evolution. SPEA and MIGA are assigned a population size of 30 and 70 for their external/archive and evolving population size, respectively, which form an overall population size of 100. All approaches under comparison were implemented with the same common sub-functions using the same programming language in Matlab (The Math Works, 1998) on an Intel Pentium II 450 MHz computer. Each simulation is terminated automatically when a fixed simulation period of 180 seconds is reached. The simulation period is determined, after a few preliminary runs, in such a way that different performance among the algorithms could be observed. To avoid random effects, 30 independent simulation runs, with randomly initialized population, have been performed on each algorithm and the performance distributions are visualized in the box plot format (Chambers et al. 1983; Zitzler & Thiele, 1999).

Figure 19 displays the performance of *SSC* (size of space covered) for each algorithm. In general, SPEA and MOEA produce a relatively high value of *SSC* indicating their ability to have a more distributed discovered Pareto-optimal front and/or to produce more non-dominated solutions that are nearer to the global trade-offs. It can also be observed that, compared to the others, FFGA, SPEA and MOEA are more consistent in the performance of *SSC*. The performance of *UD* (uniform distribution) for all algorithms is summarized in Figure 20. In general, the *UD* distributions are mostly overlapping with each other and thus there is too little evidence to draw any strong conclusion. However, as the average performance is concerned (see bold horizontal line in the box plots), SPEA, MIGA and MOEA outperform others slightly and are more consistent in terms of the measure of *UD*. Figure 21 shows the distribution of *Neval* (number of function evaluation) performed by each algorithm in a specified time. More function evaluations in a fixed CPU time indirectly indicates that less CPU time is required by the algorithm. Intuitively, this means less computational efforts are required by the algorithm to find the trade-offs. As shown in Figure 21, MIGA requires the least algorithm effort while the performances of FFGA, NPGA and MOEA are moderate in terms of *Neval*. It can also be observed that SPEA and NGGA are suitable for problems with time-consuming function evaluations: the effects in algorithm effort become less significant in these problems. In summary, the results show that MOEA requires moderate computational effort and exhibits a relatively good performance in terms of *SSC* and *UD* on the test problem, as compared to other MO evolutionary optimization methods in this study.

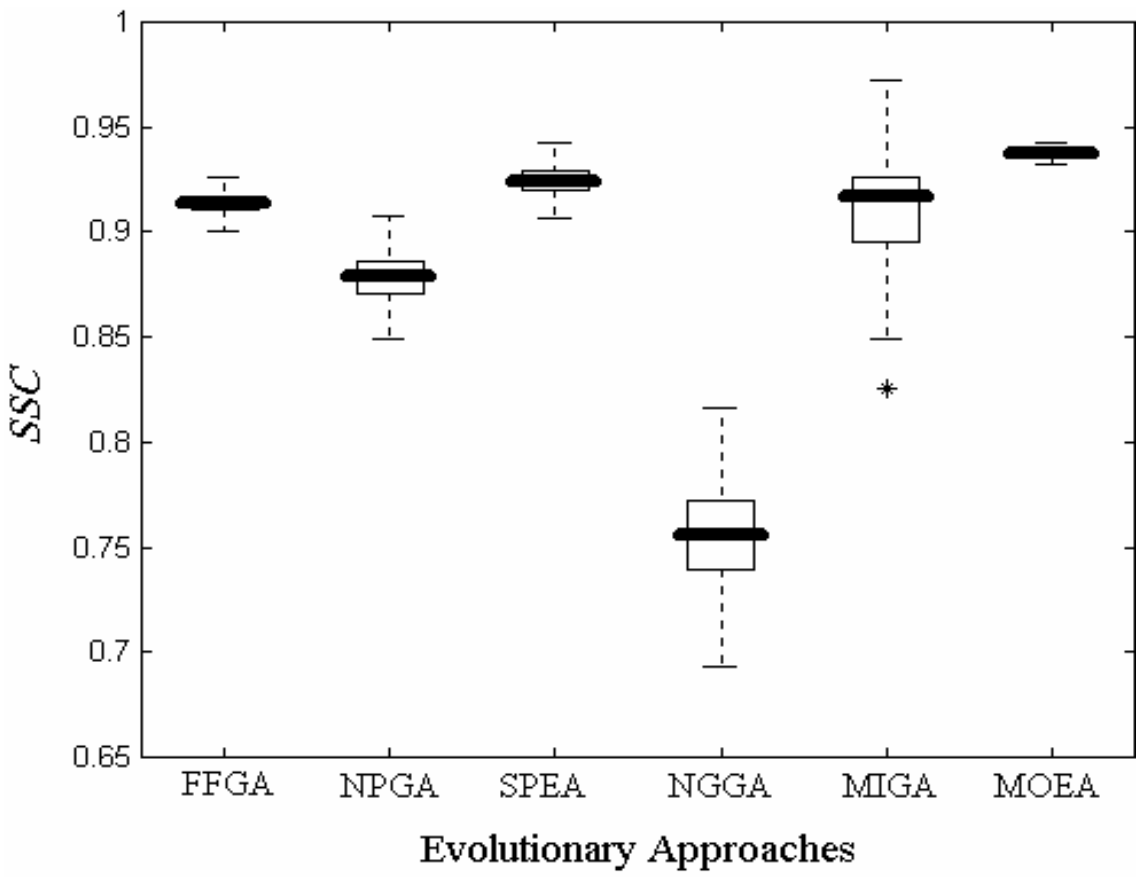

Figure 19: Box Plot of *SSC*

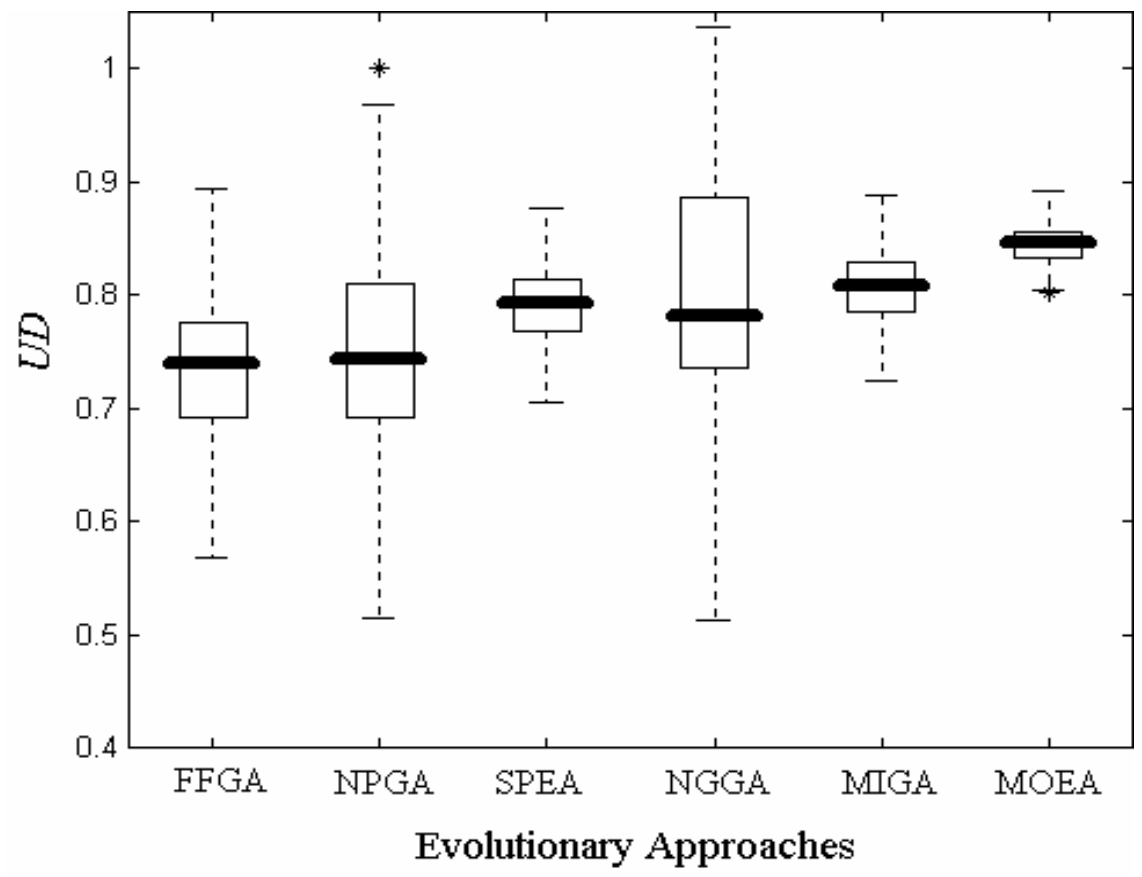

Figure 20: Box Plot of *UD*

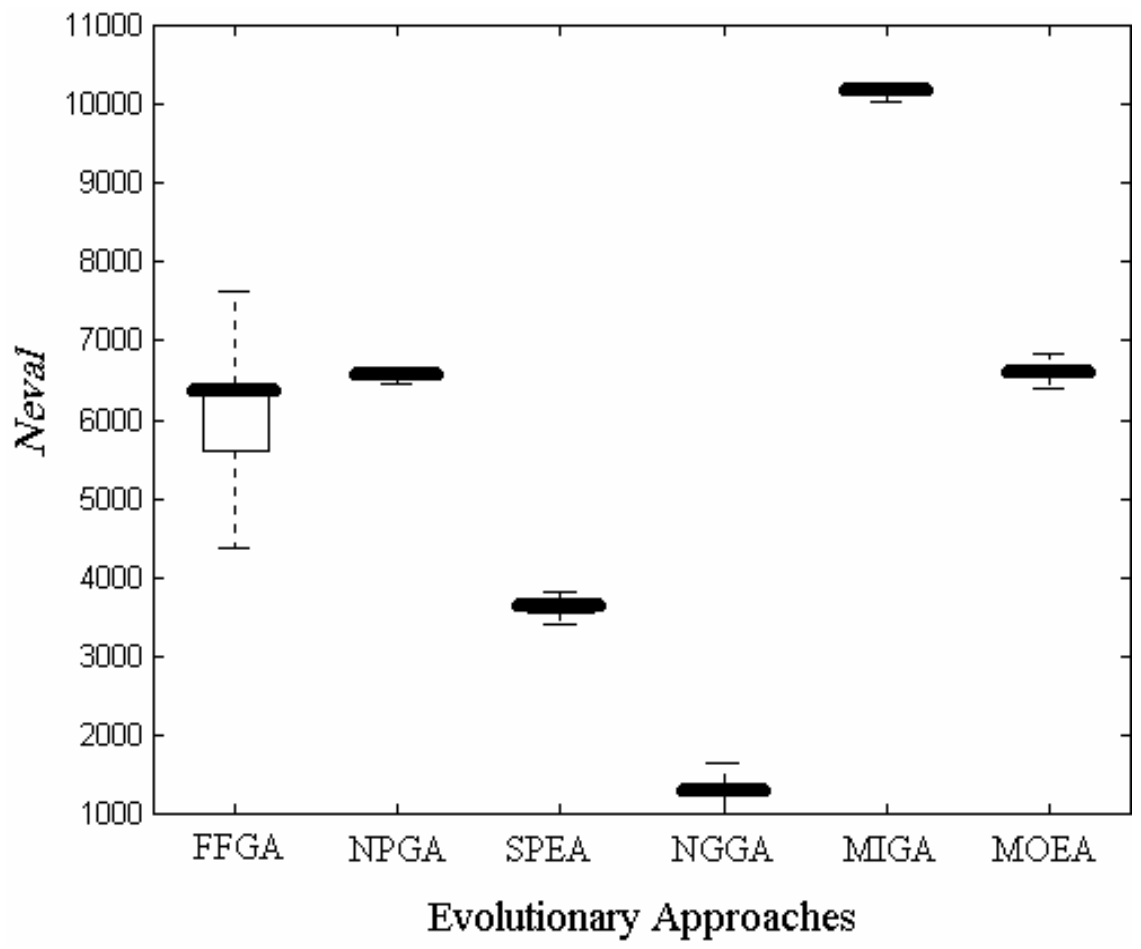

Figure 21: Box Plot of *Neval*

Figure 22 shows the distribution of non-dominated individuals in the objective domain, where the range of each axis is identical to the range shown in Figure 18. For each algorithm, the distribution is the best selected, among the 30 independent runs, with respect to the measure of *SSC*. It can be seen from Figure 22 that MOEA benefits from evolving more non-dominated individuals than the other methods. MOEA's individuals are also better distributed within the trade-off region.

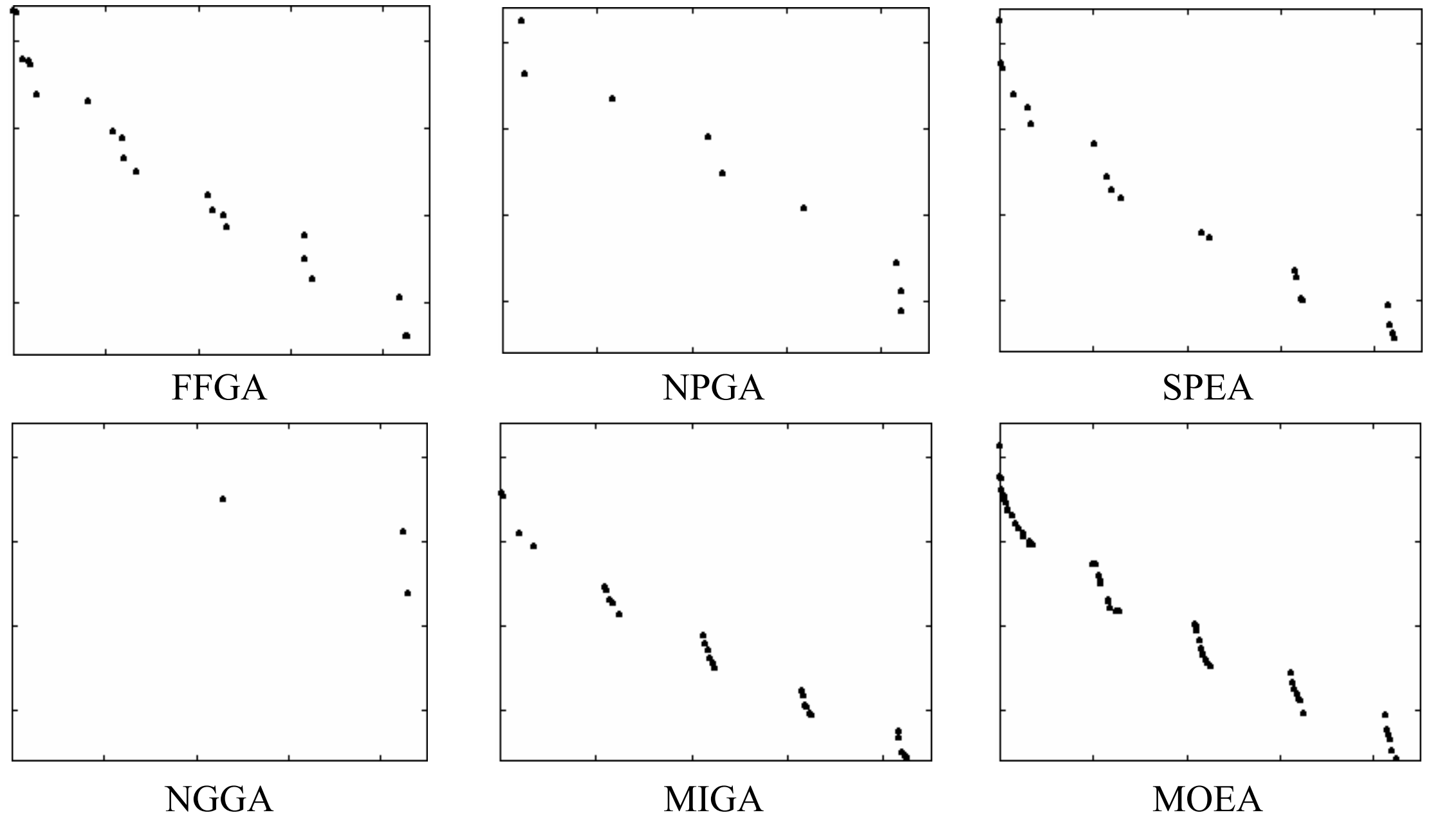

Figure 22: Best Selected Distribution of Non-dominated Individuals from Each Algorithm with Respect to the Measure of *SSC*

## 5. Application to Practical Servo Control System Design

### 5.1 The Hard Disk Drive Servo System

A typical plant model of hard disk drive (HDD) servo system includes a driver (power amplifier), a VCM (Voice Coil Motor) and a rotary actuator that is driven by the VCM. Figure 23 (Goh et al. 2001) shows a basic schematic diagram of a head disk assembly (HDA), where several rotating disks are stacked on the spindle motor shaft.

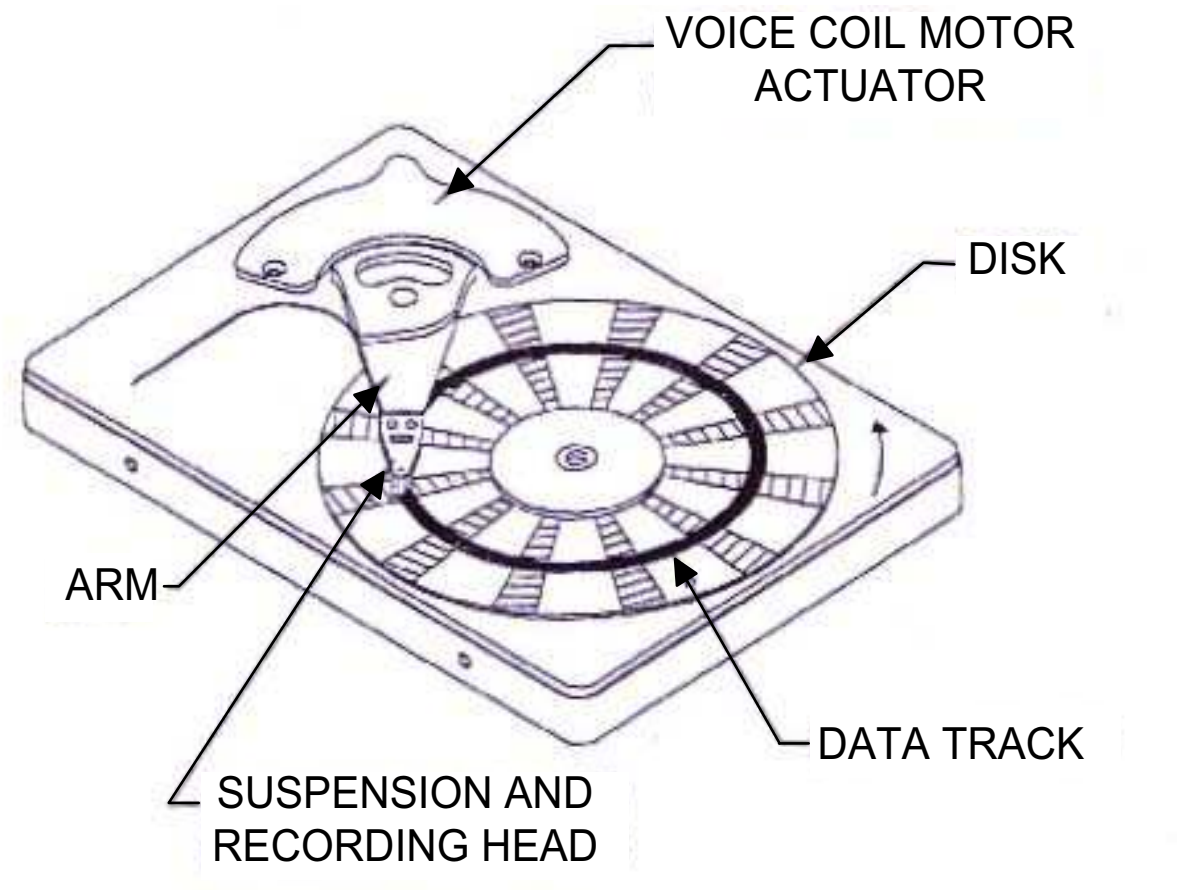

Figure 23: A HDD with a Single VCM Actuator Servo System

The dynamics of an ideal VCM actuator is often formulated as a second-order state-space model (Weerasooriya, 1996),

$$\begin{pmatrix} \dot{y} \\ \dot{v} \end{pmatrix} = \begin{pmatrix} 0 & K_y \\ 0 & 0 \end{pmatrix} \begin{pmatrix} y \\ v \end{pmatrix} + \begin{pmatrix} 0 \\ K_v \end{pmatrix} u \tag{15}$$

where $u$ is the actuator input (in volts), $y$ and $v$ are the position (in tracks) and the velocity of the R/W head, $K_v$ is the acceleration constant and $K_y$ the position measurement gain, where $K_y = K_t / m$ with $K_t$ being the current-force conversion coefficient and $m$ being the mass of the VCM actuator. The discrete-time HDD plant model used for the evolutionary servo controller design in this study is given as (Tan et al. 2000),

$$x(k+1) = \begin{pmatrix} 1 & 1.664 \\ 0 & 1 \end{pmatrix} x(k) + \begin{pmatrix} 1.384 \\ 1.664 \end{pmatrix} u \tag{16}$$

### 5.2 Evolutionary HDD Controller Design and Implementation

A two-degree-of-freedom (2DOF) control structure is adopted for the read/write head servo system as shown in Figure 24. For simplicity and easy implementation, a simple first-order discrete-time controller with a sampling frequency of 4 kHz is used for the feedforward and feedback controllers, which is in the form of

$$K_p = K_f\left(\frac{z + ff_1}{z + ff_2}\right) \qquad\qquad K_s = K_b\left(\frac{z + fb_1}{z + fb_2}\right) \tag{17}$$

respectively. The control objective during the tracking in HDD is to follow the destination track with a minimum tracking error. Note that only time domain performance specifications are considered in this paper, and the design task is to search for a set of optimal controller parameters $\{K_f, K_b, ff_1, ff_2, fb_1, fb_2\}$ such that the HDD servo system meets all design requirements. These requirements are that overshoots and undershoots of the step response should be kept less than 5% since the head can only read or write within ±5% of the target; the 5% settling time in the step response should be less than 2 milliseconds and settle to the steady-state as quickly as possible (Goh et al. 2001). Besides these performance specifications, the system is also subject to the hard constraint of actuator saturation, i.e., the control input should not exceed ±2 volts due to the physical constraint on the VCM actuator.

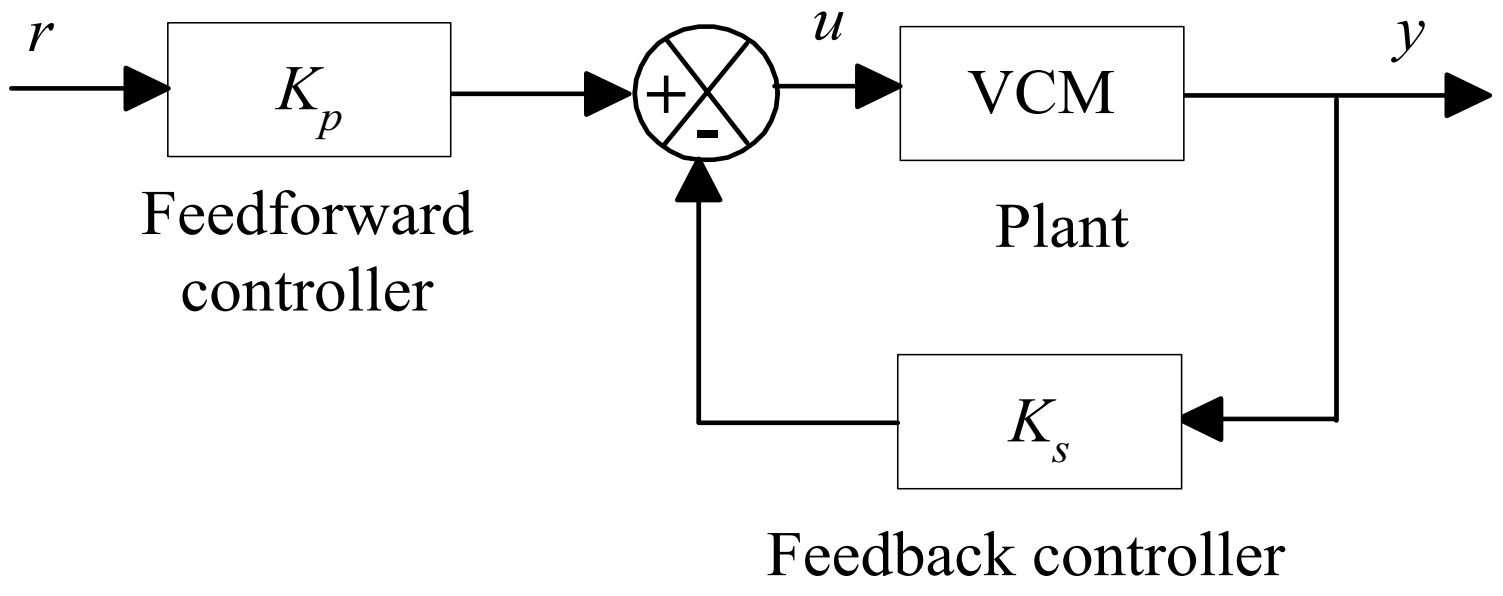

Figure 24: The Two Degree-of-freedom Servo Control System

The multi-objective evolutionary algorithm (MOEA) proposed in this paper has been embedded into a powerful GUI-based MOEA toolbox (Tan et al. 2001b) for ease-of-use and for straightforward application to practical problems. The toolbox is developed under the Matlab (The Math Works, 1998) programming environment, which allows users to make use of the versatile Matlab functions and other useful toolboxes such as Simulink (The Math Works, 1999). It allows any trade-off scenario for MO design optimization to be examined effectively, aiding decision-making for a global solution that best meets all design specifications. In addition, the toolbox is equipped with a powerful graphical user interface (GUI) and is ready for immediate use without much knowledge of evolutionary computing or programming in Matlab. A file handling capability for saving all simulation results and model files in a Mat-file format for Matlab or text-file format for software packages like Microsoft Excel is also available in the toolbox. Through the GUI window of MOEA toolbox, the time domain design specifications can be conveniently set as depicted in Figure 25, where *Tr*, *OS*, *Ts*, *SSE*, *u* and *ue* represents the rise time, overshoot, settling time, steady-state error, control input and change in control input, respectively.

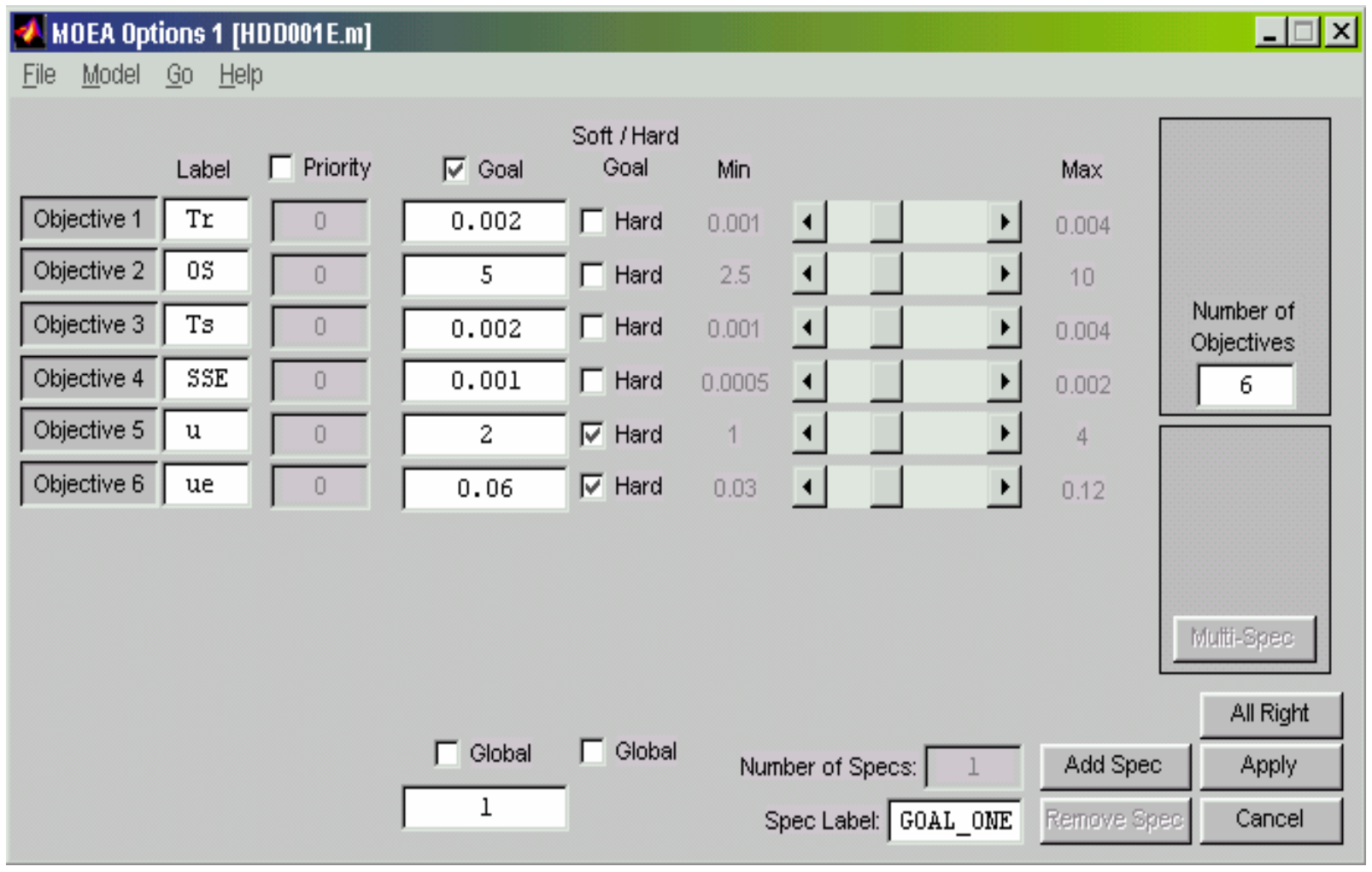

Figure 25: MOEA GUI Window for Settings of Design Specifications

The simulation adopts a generation and population size of 200, and all the design specifications listed in Figure 25 have been successfully satisfied at the end of the evolution. The design trade-off graph is shown in Figure 26, where each line representing a solution found. The *x*-axis shows the design specifications and the *y*-axis shows the normalized cost for each objective. Clearly, trade-offs between adjacent specifications result in the crossing of the lines between them (e.g., steady-state error (*SSE*) and control effort (*u*)), whereas concurrent lines that do not cross each other indicating the specifications do not compete with one another (e.g., overshoots (*OS*) and settling time (*Ts*)).

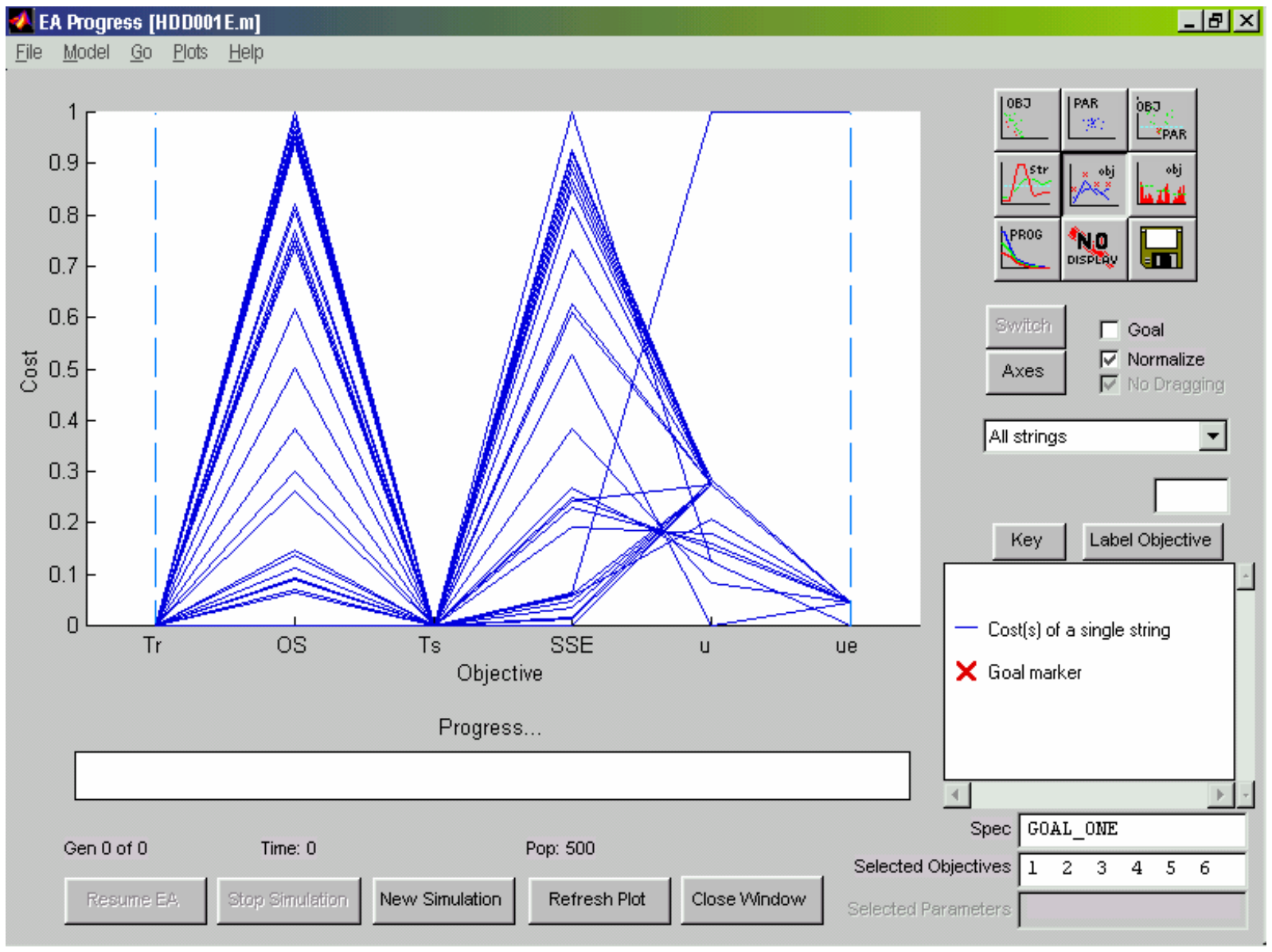

Figure 26: Trade-off Graph of the HDD Servo Control System Design

The closed-loop step response of the overall system for an arbitrary selected set of MOEA designed 2DOF controller parameters given as $\{K_f, K_b, ff_1, ff_2, fb_1, fb_2\}$ = {0.029695, -0.58127, 0.90279, -0.3946, -0.70592, 0.83152} is shown in Figure 27. With a sampling frequency of 4 kHz, the time domain closed-loop performance of the evolutionary designed controller has been compared with the manually designed discrete-time PID controller as given in Eq. 18 (Goh et al. 2001) as well as the Robust and Perfect Tracking (RPT) controller (Goh et al. 2001) as given in Eq. 19,

$$u = \frac{0.13z^2 - 0.23z + 0.1}{z^2 - 1.25z + 0.25}(r - y) \tag{18}$$

$$\begin{aligned} x(k+1) &= -0.04x(k) + 15179r(k) - 453681y(k) \\ u(k) &= -3.43 \times 10^{-7} x(k) + 0.04r(k) - 0.18y(k) \end{aligned} \tag{19}$$

It can be seen in Figure 27 that the evolutionary designed 2DOF controller has outperformed both the PID and RPT controllers, with the fastest rise time, smallest overshoots and shortest settling time in the closed-loop response. Its control performance is excellent and the destination track crossover occurs at approximately 1.8 milliseconds.

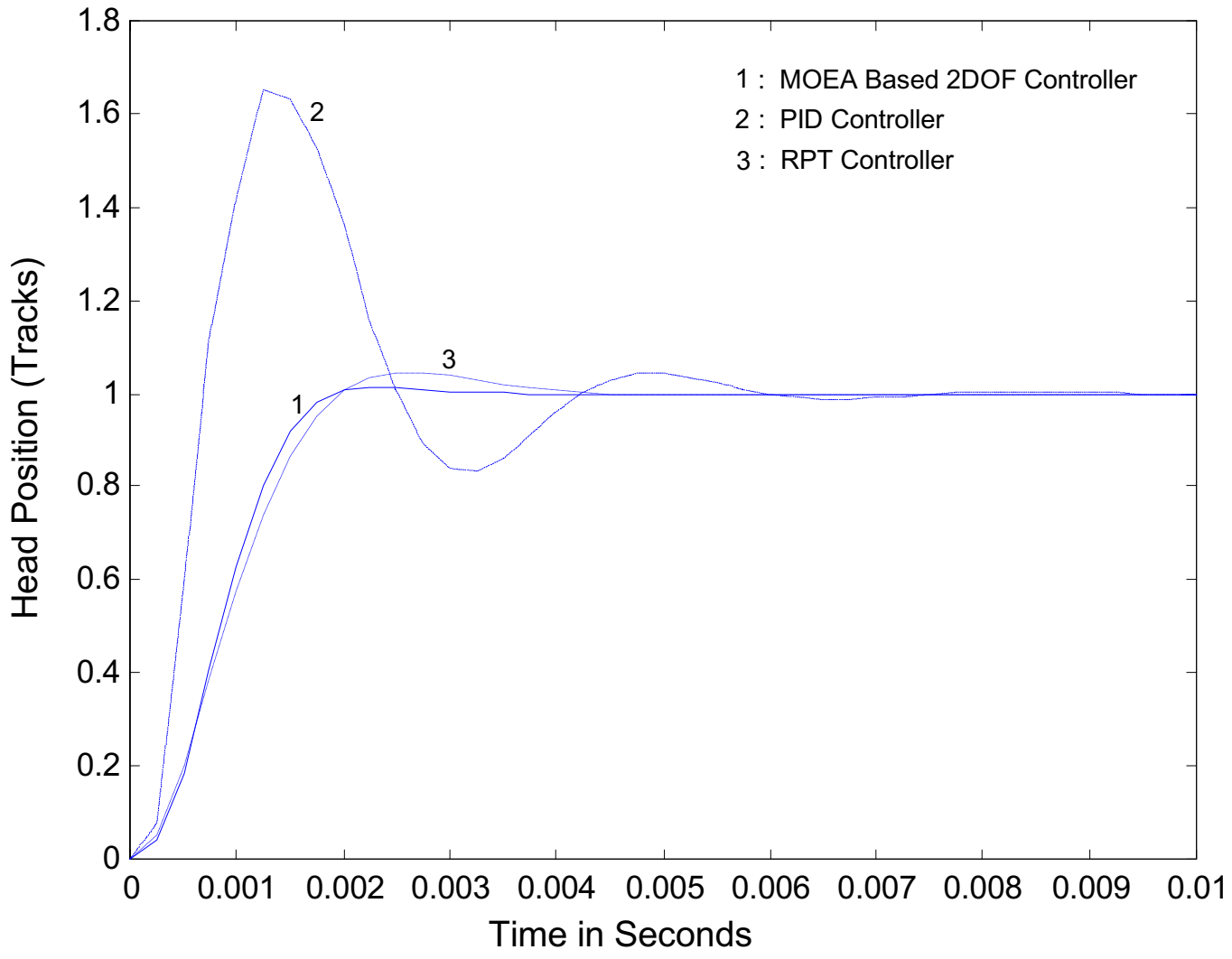

Figure 27: Closed-loop Servo System Responses with Evolutionary 2DOF, RPT and PID Controllers

The performance of the evolutionary 2DOF servo control system was further verified and tested on the physical 3.5-inch HDD with a TMS320 digital signal processor (DSP) and a sampling rate of 4 kHz. The R/W head position was measured using a laser doppler vibrometer (LDV) and the resolution used was 1 μm/volt. Real-time implementation result of the evolutionary HDD servo control system is given in Figure 28, which is consistent with the simulated step response in Figure 27, and shows an excellent closed-loop performance.

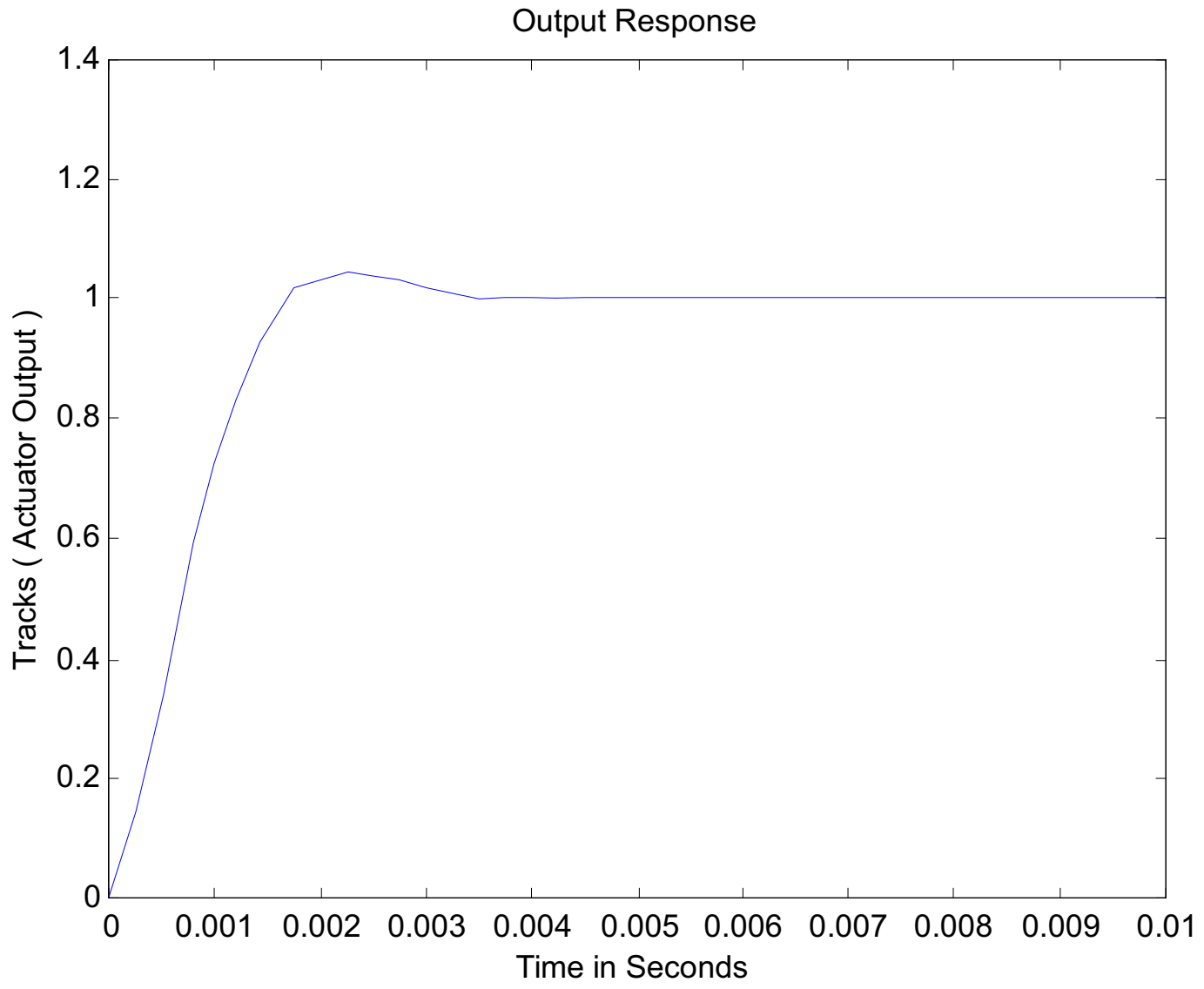

Figure 28: Real-time Implementation Response of the Evolutionary 2DOF Servo System

## 6. Conclusions

This paper has presented a multi-objective evolutionary algorithm (MOEA) with a new goal-sequence domination scheme to allow advanced specifications such as hard/soft priorities and constraints to be incorporated for better decision support in multi-objective optimization. In addition, a dynamic fitness sharing scheme that is simple in computation and adaptively based upon the on-line population distribution at each generation has been proposed. Such a dynamic sharing approach avoids the need for a priori parameter settings or user knowledge of the usually unknown trade-off surface often required in existing methods. The effectiveness of the proposed features in MOEA has been demonstrated by showing that each of the features contains its specific merits and usage that benefit the performance of MOEA. In comparison with other existing evolutionary approaches, simulation results show that MOEA has performed well in the diversity of evolutionary search and uniform distribution of non-dominated individuals along the final trade-offs, without significant computational effort. The MOEA has been applied to the practical engineering design problem of a HDD servo control system. Simulation and real-time implementation results show that the evolutionary designed servo system provides excellent closed-loop transient and tracking performance.

## Acknowledgements

The authors wish to thank Andrew Moore and the anonymous reviewers for their valuable comments and helpful suggestions which greatly improved the paper quality.